\documentclass[lettersize,journal]{IEEEtran}

\usepackage{url}
\usepackage{bbm}
\usepackage{cite}
\usepackage{caption}
\usepackage{xcolor}
\usepackage{xspace}
\usepackage{gensymb}
\usepackage{booktabs}
\usepackage{multirow}
\usepackage{pifont}
\usepackage{balance}
\usepackage{graphicx}
\usepackage{subcaption}
\usepackage[export]{adjustbox}
\usepackage[noend]{algpseudocode}
\usepackage{amsmath,amssymb,amsfonts}
\usepackage[linesnumbered,ruled]{algorithm2e}
\def\S{\emph{FacER+}\xspace}

\def\etal{\textit{et al.}\xspace}

\def\eg{\textit{e.g.}\xspace}

\def\iid{\textit{i.i.d.}\xspace}

\DeclareMathOperator*{\argmin}{arg\,min}

\begin{document}

\title{Decoding Emotions: Unveiling Facial Expressions through Acoustic Sensing with Contrastive Attention}


\author{
\IEEEauthorblockN{Guangjing Wang, Juexing Wang, Ce Zhou, Weikang Ding, Huacheng Zeng, Tianxing Li and Qiben Yan}\\
\IEEEauthorblockA{Department of Computer Science and Engineering, Michigan State University, USA}\\
\{wanggu22, wangjuex, zhouce, dingweik, hzeng, litianx2, and qyan\}@msu.edu
}

\markboth{IEEE Transactions on Mobile Computing, ~Vol.~1, No.~1, December~2024}%
{Shell \MakeLowercase{\textit{et al.}}: A Sample Article Using IEEEtran.cls for IEEE Journals}


\maketitle


\label{abstract}
\begin{abstract}
Expression recognition holds great promise for applications such as content recommendation and mental healthcare by accurately detecting users' emotional states. Traditional methods often rely on cameras or wearable sensors, which raise privacy concerns and add extra device burdens. In addition, existing acoustic-based methods struggle to maintain satisfactory performance when there is a distribution shift between the training dataset and the inference dataset. In this paper, we introduce FacER+, an active acoustic facial expression recognition system, which eliminates the requirement for external microphone arrays. FacER+ extracts facial expression features by analyzing the echoes of near-ultrasound signals emitted between the 3D facial contour and the earpiece speaker on a smartphone. This approach not only reduces background noise but also enables the identification of different expressions from various users with minimal training data. We develop a contrastive external attention-based model to consistently learn expression features across different users, reducing the distribution differences. Extensive experiments involving 20 volunteers, both with and without masks, demonstrate that FacER+ can accurately recognize six common facial expressions with over 90\% accuracy in diverse, user-independent real-life scenarios, surpassing the performance of the leading acoustic sensing methods by 10\%. FacER+ offers a robust and practical solution for facial expression recognition. The source code is available at~\url{https://github.com/MyRespect/FaceAcousticSensing}.
\end{abstract}


\section{Introduction}
\label{introduction}
In the mobile-centric digital era, various social media platforms such as YouTube, Facebook, and TikTok compete for user engagement through smartphones and other mobile devices. Understanding fine-grained emotional responses is crucial for enhancing user interactions with social media platforms. Traditionally, user feedback on services has been gauged through crowd-sourced ratings and reviews, which lack the granularity for capturing real-time, spontaneous reactions. To deliver more tailored experiences, it is essential to develop a precise and reliable method for detecting users' emotions and gathering their immediate feedback.

Numerous methods have been proposed for emotion recognition, utilizing diverse biometric indicators including facial expressions~\cite{chen2021exgsense, gao2021sonicface, krumhuber2023role}, vocal characteristics~\cite{nwe2001speech, swain2018databases, wani2021comprehensive}, and cardiac rhythms~\cite{zhao2016emotion, saganowski2022emotion, zhang2020emotion}.
Nevertheless, facial expressions are widely regarded as the most straightforward method for decoding human emotions, serving as a universal medium of nonverbal communication~\cite{li2020deep}. 
The Facial Action Coding System (FACS)~\cite{ekmann1973universal} models six commonly recognized facial expressions (FEs): anger, disgust, fear, happiness, sadness, and surprise. FACS comprises action units (basic facial muscles involved) and action descriptors (singular movements of multiple muscle groups)~\cite{clark2020facial}. When individuals display various facial expressions, distinct facial muscles are activated, which can be detected through diverse sensing signals.

Current facial expression recognition (FER) techniques are classified into three main categories: camera-based~\cite{cao2013facewarehouse, patil2016real, shen2016facial}, radio-based~\cite{gu2020wife, chen2020wiface, choi2022ppgface}, and acoustic-based expression recognition~\cite{gao2021sonicface, wang2023facer, wang2024uface}. 
However, camera-based methods for facial expression recognition, such as FaceWarehouse~\cite{cao2013facewarehouse} which collects RGBD data from various users, raise significant privacy concerns due to their reliance on continuous video recording. These concerns hinder their widespread adoption in real-world applications. Additionally, these methods struggle to accurately recognize facial expressions when users' faces are occluded, further limiting their effectiveness. 
Other FER methods often depend on additional hardware, which can hardly be adopted in real life. For instance, WiFace~\cite{chen2020wiface} utilizes a WiFi router equipped with three antennas in fixed positions for FER, necessitating extra hardware and setup. Similarly, PPGface~\cite{choi2022ppgface} relies on costly wearable devices equipped with photoplethysmography sensors, increasing the barrier to widespread use. Another limitation of current FER methods is their inability to generalize effectively to new users. SonicFace~\cite{gao2021sonicface} and UFace~\cite{wang2024uface} are both acoustic-based facial expression recognition methods, but they struggle to manage variations in expressions that are not included in the training dataset. For example, the leave-one-user-out accuracy for UFace is only 61.65\%. Additionally, labeled data from new users is required to achieve user adaptation, but it is hard and inconvenient to obtain labeled data from new users. 

\begin{figure}[t]
\centering
\begin{subfigure}{0.225\textwidth}
\centering
    \includegraphics[width=0.8\linewidth]{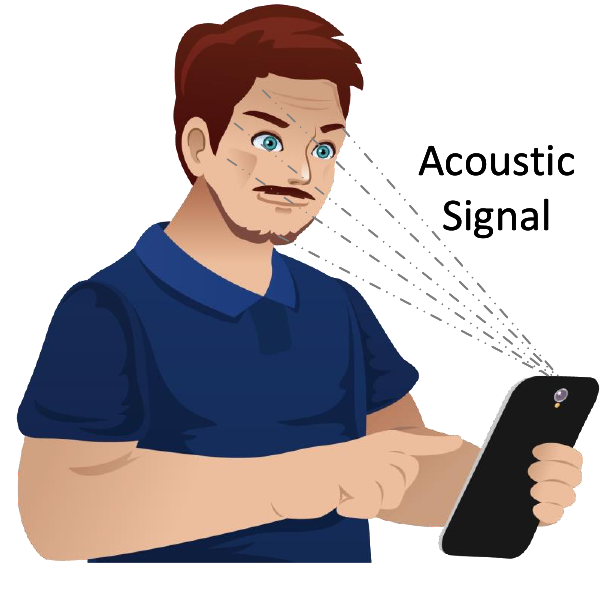}
    \caption{FacER application scenario.}
    \label{fig:scenario}
\end{subfigure}
\begin{subfigure}{0.24\textwidth}
\centering
  \includegraphics[width=0.915\linewidth]{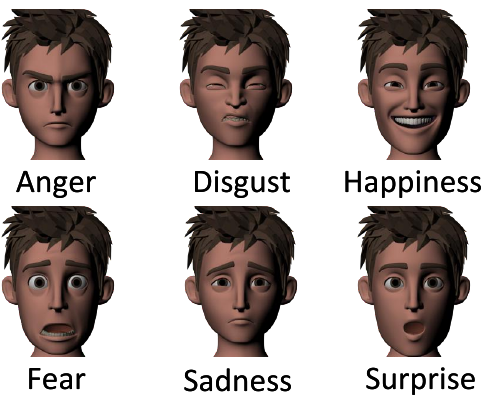}
  \caption{Six facial expressions~\cite{aneja2016modeling}.}
  \label{fig:fe}
\end{subfigure}
\caption {Facial expression recognition using a smartphone.}
\label{fig:scenario_fe}
\end{figure}

In this work, we introduce \S, a \textbf{Fac}ial \textbf{E}xpression \textbf{R}ecognition system that employs near-ultrasound acoustic sensing. Our approach utilizes a commercial smartphone to emit near-ultrasound signals (19-23 kHz) directed toward the user's face, without requiring additional hardware such as microphone array~\cite{gao2021sonicface} or WiFi router~\cite{chen2020wiface}. As depicted in Fig.~\ref{fig:scenario}, \S utilizes the microphones in a smartphone to capture echoes reflected off the user's face. These echoes reflect facial muscle movement patterns about expressions. By analyzing the intricate patterns of these echoes, \S can distinguish six different types of facial expressions, as illustrated in Fig.\ref{fig:fe}. In addition, \S can overcome the facial occlusion issues encountered by camera-based methods. Our experiments demonstrate that \S can recognize six universal facial expressions with an accuracy exceeding 85\% in facial occlusion scenarios (\eg, wearing a face mask). Furthermore, \S achieves high-performance user adaptation without requiring data labels, avoiding the burden of labeling data. There are three main challenges in designing \S.

First, acoustic noise, including signal reflection multipaths and ambient noise, can greatly affect expression recognition accuracy. Besides capturing echoes from the face, microphones also receive reflections from nearby objects, complicating the signal analysis. It is crucial to eliminate the impact of noise in received signals. Despite employing various noise-cancellation techniques as in~\cite{wang2024uface}, environmental noises that share a similar frequency with the emitted signal persist, degrading model performance. Therefore, besides the noise filtering methods, we design an external cross-sample attention-based learning model to capture robust expression feature representations. This approach enables the model to extract the essential features of expressions while filtering out background noises.

Second, the expression sensing data collection and labeling cost heavily. High-quality data collection demands a controlled setting, which makes it challenging for individuals to maintain facial expressions for data collection. In addition, the process of collecting and labeling data is time-intensive, requiring significant manual effort and coordination. Therefore, we design a series of data augmentation methods including inter and intra-sample augmentation to enlarge the training data size while reducing the data collection and labeling costs. Given the intricate temporal dynamics of acoustic-sensing data and complex facial expressions, we must make sure the ad-hoc transformation does not change the fine-grained expression features in the acoustic signal.

Third, the way individuals show facial expressions varies. For the same expression, the same user potentially expresses them differently at different times. This variation leads to the challenge of domain adaptation, a prevalent issue in machine learning (ML) applications. Typically, an ML model trained on a labeled dataset (source domain) struggles to perform effectively on a different testing dataset (target domain) due to distribution drift between the two domains. This drift violates the standard independent and identically distributed (\iid) assumption underlying most ML models. Therefore, we design a domain adaptation contrastive learning algorithm to align the distributions of the source and target domain datasets, as well as the synthetic and real domain datasets. This approach enables consistent performance in recognizing a range of expressions across different unseen users.

We evaluate \S on a dataset collected from 20 volunteers across a two-year period, varying in age, gender, and skin color, across diverse environments and different times. The results demonstrate that \S can effectively recognize six distinct facial expressions from various users. Impressively, it achieves more than 97\% accuracy when the training and testing datasets are similarly distributed. Even when trained and tested on different user groups, it maintains an accuracy of over 90\%. In summary, our contributions are as follows:

\begin{itemize}
\item We develop \S, an acoustic facial expression recognition model that leverages contrastive external attention to capture distinctive and robust facial expression features, simultaneously filtering out background noise.

\item We design data augmentation methods and a domain adaptation algorithm to synchronize the distributions of training and testing data, and distributions of synthetic and real data, effectively reducing the impact of variability in users' facial expressions.

\item We implement the smartphone-based system \S, and conduct tests in various real-world conditions. Our findings reveal that \S surpasses existing methods, improving recognition accuracy by over 10\% while offering enhanced mobility and user convenience.
\end{itemize}

The rest of the paper is organized as follows. In Section~\ref{related_work}, we summarize the related work. We introduce the preliminary knowledge in Section~\ref{preliminary}. 
In Section~\ref{system}, we present \S and the proposed data augmentation method and the contrastive attention model. We provide implementation details in Section~\ref{implementation} and evaluate the performance of \S in Section~\ref{evaluation}. We discuss the future work in Section~\ref{discussion}.
Finally, we conclude in Section~\ref{conclusion}.

\section{Related Work}
\label{related_work}
To recognize the emotional states of users, researchers have proposed to use body sensors to monitor physiological information such as electromyographic (EMG) signals~\cite{chen2021exgsense, gruebler2014design} and heart rate~\cite{rommel2012heart, qian2018acousticcardiogram}. Yet, this method often demands a lengthy analysis period, such as 30 seconds~\cite{fleureau2013affective}, to effectively profile emotions, leading to inefficiency. ExpressEar~\cite{verma2021expressear} integrates commercial earables with inertial sensors to detect movements in facial muscles related to expressions. NeckFace~\cite{chen2021neckface} utilizes a neckpiece equipped with infrared (IR) cameras to monitor facial expressions. Similarly, FaceListener~\cite{songfacelistener2022} uses headphones converted into acoustic sensors to track deformations in facial skin as a means of recognizing expressions. Nonetheless, the necessity for users to wear these devices can be inconvenient and restrictive.

Another method, wireless and mobile sensing, has been employed for behavior recognition tasks such as identifying daily activities~\cite{wang2018socialite, han2019shad, zheng2019zero, li2020wihf, li2021deep} and facial expressions~\cite{chen2020wiface, rostaminia2019w}. For example, WiFace~\cite{chen2020wiface} utilizes the channel state information in WiFi signals captured by a router equipped with three antennas positioned above the user’s head. The resulting waveform patterns are used to recognize facial expressions. Similarly, Hof \etal~\cite{hof2020face} introduce a mm-wave radar system with a large number of antenna elements for facial recognition. Nevertheless, these methods necessitate additional hardware and specific placement configurations.

Given the widespread availability of speakers and microphones, acoustic sensing has been extensively studied. The core concept involves using a speaker to emit acoustic signals and analyzing the echoes reflected by sensing objects. This technique has a broad range of applications, including breath monitoring~\cite{xu2019breathlistener}, user authentication~\cite{zhou2018echoprint}, and activity recognition and tracking~\cite{gao2022mom, xie2022teethpass, li2022lasense, mao2019rnn, gao2020echowhisper, zhang2021soundlip, iravantchi2019beamband, li2022eario}. For instance, EchoPrint~\cite{zhou2018echoprint} integrates acoustic and visual signals for user authentication by emitting inaudible acoustic signals toward the user's face and extracting features from the echoes reflecting off the 3D facial contour. Similarly, TeethPass~\cite{xie2022teethpass} employs earbuds to capture occlusal sounds in binaural canals for user authentication. Zhang~\etal\cite{zhang2024face} analyze the acoustic signal reflected by the human face and generate facial spectrums for face recognition, achieving more than 95\% recognition accuracy. LASense~\cite{li2022lasense} accomplishes fine-grained activity sensing by increasing the number of overlapped samples between the emitted and received acoustic signals through signal processing, thereby enhancing both sensing accuracy and range. 

Numerous acoustic-based expression sensing techniques~\cite{xie2021acoustic, wang2024uface, gao2021sonicface, wang2023facer} have been proposed. For instance, SonicFace~\cite{gao2021sonicface} detects expressions using a customized microphone array to capture reflected echoes. It calculates the frequency and phase shifts of pure tone signals to extract expression features. However, SonicFace requires additional hardware in the fixed position. Also, the expression recognition performance is limited due to noise interference. Later, our previous work Facer~\cite{wang2023facer} is designed to use the earpiece speaker on a smartphone to build an expression recognition model that adapts to different users. Then, UFace~\cite{wang2024uface} improves the robustness of the expression recognition system on smartphones by considering more conditions to verify the front face state and eliminate self-interference, such as finger swipes. However, both SonicFace and UFace suffer from the domain gap issue~\cite{wei2018person} caused by the differences between the training dataset and inference dataset. Hence, they only achieve limited performance to unseen users and environments.

Substantial efforts have been spent on mitigating domain gaps by designing cross-domain adaptation solutions. For example, Widar3.0~\cite{zheng2019zero} is a cross-domain gesture recognition system via Wi-Fi, which estimates velocity profiles of gestures at the signal level. However, the tiny movement of muscles on the face makes their velocity profiles hard to estimate. XHAR~\cite{zhou2020xhar} is a domain adaptation framework for activity recognition based on adversarial training. However, it is hard to accurately define domain relevance and achieve the adversarial optimization objectives.
Moreover, the domain adaptation methods are usually designed based on the features of the data and the specific task. In this work, we design a contrastive attention domain adaptation method based on the features of both acoustic signals and facial expressions to reduce the discrepancy between different domains and enhance the model's performance.

\section{Preliminaries}
\label{preliminary}
In this section, we present the background of acoustic signals, an overview of attention mechanisms, and the foundational concepts of contrastive learning.

\begin{figure}[t]
\begin{subfigure}{0.225\textwidth}
    \includegraphics[width=\linewidth]{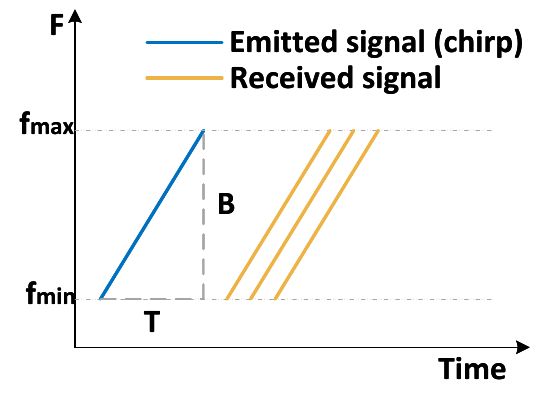}
    \caption{Illustration of FMCW.}
    \label{fig:fmcw}
\end{subfigure}
\begin{subfigure}{0.225\textwidth}
  \includegraphics[width=0.98\linewidth]{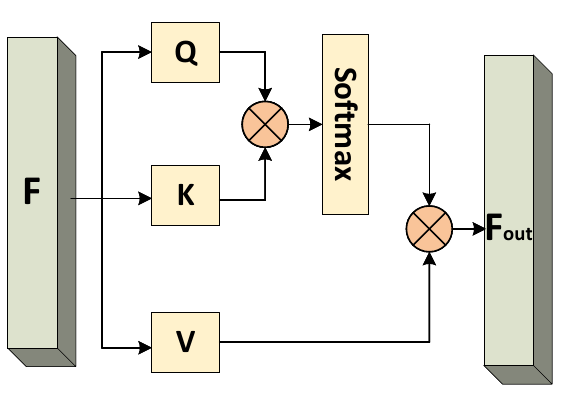}
  \caption{Illustration of self-attention.}
  \label{fig:attention}
\end{subfigure}
\caption {Illustration of Preliminaries.}
\label{fig:preliminary}
\vspace{-10pt}
\end{figure}

\subsection{Acoustic Signal}
The acoustic signal refers to a coded chirp signal transmitted by a device. Specifically, a chirp sound signal is a frequency-swept signal, modulated in frequency, as illustrated in Fig.~\ref{fig:fmcw}. The chirp signal can be considered a component of sawtooth modulation in Frequency-Modulated Continuous Wave (FMCW), where the operating frequency changes during measurement. In FMCW, the signal's frequency periodically increases or decreases during transmission. The frequency differences between the transmitted and received signals are proportional to the time delay $\Delta t$. Consequently, FMCW can measure the small movements of the target, which is calculated as follows:
\begin{equation}
    R=\frac{v_0|\Delta t|}{2}=\frac{v_0|\Delta f|T}{2 B},
    \label{eq:fmcw}
\end{equation}
where $R$ represents the distance between the sound source and the reflecting object, $v_0$ is the speed of sound (340 m/s) at 20~\degree C, $\Delta t$ is the delay time, and $\Delta f$ is the measured frequency difference. $B$ denotes the chirp frequency bandwidth, and $T$ is the chirp periodic time. The duration of the transmitted waveform $T$ must exceed the required receiving time for the distance measuring range. We use $\frac{B}{T}$ to quantify the frequency shift per unit of time. Therefore, with the characteristics of FMCW, the chirp signal can group reflections from various distances into multiple range bins.

\subsection{Attention Mechanism}
Similar to the human visual system, attention mechanisms~\cite{vaswani2017attention} are designed to focus limited attention on key information, conserving resources and distilling essential data. The fundamental concept of attention mechanisms is to combine all encoded input features in a weighted manner, giving the highest weights to the most important features. The self-attention mechanism enhances the representation at each position by integrating features from other positions within a sample (e.g., an image), thereby capturing long-range dependencies.

As illustrated in Fig.~\ref{fig:attention}, given a feature map $F \in \mathcal{R}^{N\times d}$, where $N$ is the number of elements and $d$ is the feature dimension of each element, by multiplying three different random initialized weight matrixes,
self-attention projects the $F$ into a query matrix $Q\in \mathcal{R}^{N\times d'}$, a key matrix $K\in \mathcal{R}^{N\times d'}$, and a value matrix $V\in \mathcal{R}^{N\times d}$ as follows:
\begin{equation}
    F_{out} = softmax(QK^T)V,
    \label{eq:attention}
\end{equation}
where $softmax(QK^T)$ is the attention matrix, and the $F_{out}$ is the improved feature representation of the input $F$. The facial expression echos contain multiple facial muscle movements, as well as the background noise. Therefore, the neural network needs to capture different important aspects of expressions.

\subsection{Contrastive Learning}
Contrastive representation learning aims to learn an embedding space where dissimilar samples are spread out and similar samples remain close together. Normally, a positive pair refers to a pair of samples that have the same label, and a negative sample pair has different labels.

The supervised contrastive loss~\cite{khosla2020supervised} is defined as follows when the training objective includes multiple positive and negative pairs in one batch:
\begin{equation}
\begin{split}
        \mathcal{L}_\text{c} = \sum_{i\in I}\frac{-1}{|P(i)|}\sum_{p\in P(i)}log\frac{exp(z_i\cdot z_p/\tau)}{\sum_{a\in A(i)} exp(z_i\cdot z_a/\tau)},
\end{split}
\label{eq:contrastive2}
\end{equation}
where $I$ is a set of samples $x$ within a batch, $A(i) \equiv I\backslash \{i\}$, $P(i)$ is a set of indices of all positives in the multiviewed batch, $|P(i)|$ is the cardinality, $z = Proj(Enc(x))$ is the encoded feature representation by an encoder network $Enc(\cdot)$ and a projection network $Proj(\cdot)$ such as a linear layer network. The $\cdot$ denotes the inner product, and $\tau$ is a temperature parameter to adjust the final results.

\section{System Design}
\label{system}
In this section, we develop an acoustic sensing signal pre-processing method, identify the domain gap for expression recognition, and propose a data augmentation algorithm to prepare the data for expression recognition. Finally, we introduce a domain adaptation model based on contrastive external attention, designed for recognizing acoustic facial expressions.

\subsection{Acoustic Sensing Design}
\label{subsec:signal}
A distinctive facial expression contour comprises a unique assembly of reflective surfaces, each generating a specific combination of echoes. Since objects absorb and attenuate sound waves differently, it becomes feasible to distinguish between echoes reflected from objects and those emanating from facial expressions~\cite{zhou2018echoprint}.

\subsubsection{Signal Generator}

Smartphones typically feature a primary speaker and microphone located at the bottom or back, along with an earpiece speaker and microphone at the top of the device. Given the earpiece speaker's optimal positioning to direct sound towards a user's face, as illustrated in Fig.~\ref{fig:scenario}, we choose the earpiece speaker for emitting the near-ultrasound acoustic signal. Additionally, due to the natural positioning of the hand when holding a phone, the top microphone is selected as it is minimally obstructed by the hand.

The acoustic signal must adhere to several criteria: (i) It should have a moderate duration to reduce echo overlap from various distances. (ii) The signal needs to be identifiable in the frequency domain, distinct from background noise, which predominantly falls below 8 kHz. (iii) Additionally, the signal should remain inaudible in practical environments. Thus, given that facial expression changes occur within 1 second, we opt for a 25-millisecond chirp signal that sweeps from 19-23 kHz to create the inaudible acoustic signal. The FMCW-based approach enables more accurate capture of muscle movements corresponding to various facial expressions and helps distinguish echoes from various obstructions. Following the Nyquist sampling theorem, the sampling rate is established at 48 kHz. The earpiece speaker periodically emits near-ultrasound signals, while the microphone simultaneously captures signals reflected off the face. We maintain a 50-millisecond interval between emissions to ensure all echoes from the preceding chirp are received before the next chirp is transmitted, allowing for clear separation of the chirps.

\begin{figure}[t]
\centering
\includegraphics[width=0.9\linewidth]{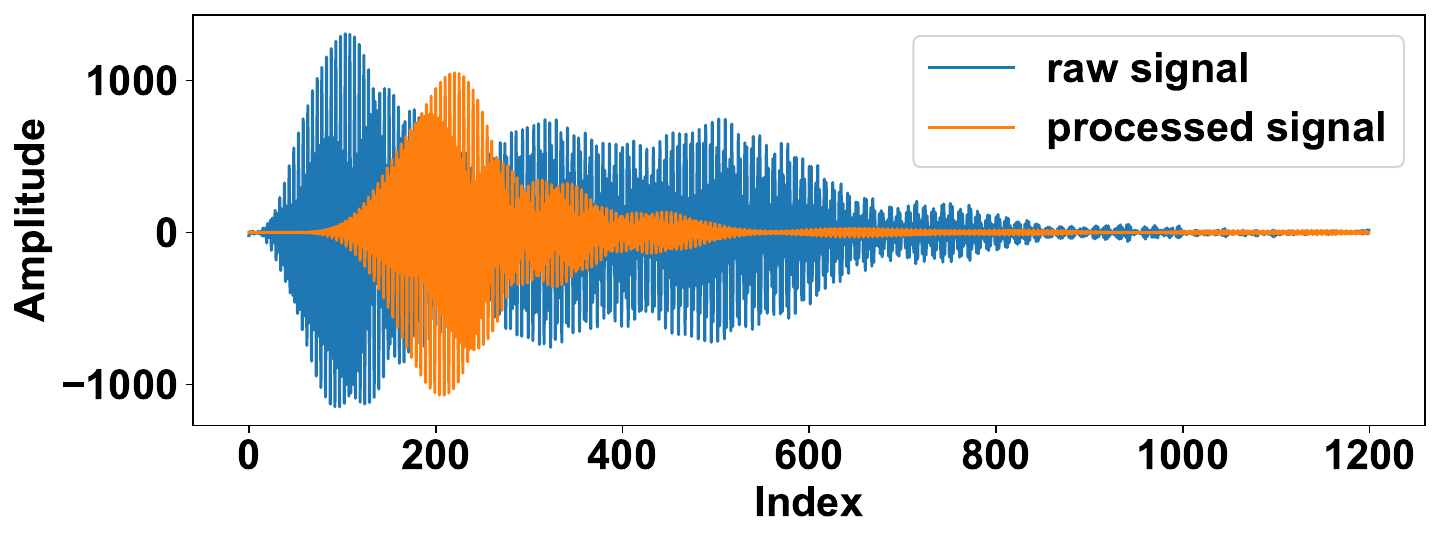}
\caption{The raw signal and the signal after noise removal.}
\label{fig:filter_signal}
\vspace{-10pt}
\end{figure}

\subsubsection{Noise Removal}
We employ a 19-23 kHz band-pass filter to eliminate expression-irrelevant signals from environmental noise. This filter retains only the desired frequency band while excluding the background noise. In addition, we adopt the disturbance removal techniques in~\cite{wang2024uface} to remove the side head orientation, finger swipe interference, as well as other facial motions such as talking and chewing. After filtering, three primary types of signals persist in the recorded output: (i) the \textit{direct path signal}, which travels straight from the speaker to the microphone; (ii) the \textit{major echo signal}, consisting of mixed echoes from the facial contour and being the focus of our study; and (iii) the \textit{noisy echo signals}, which are echoes from various obstacles within the environment due to the multiple paths of reflected signals.

To mitigate the impact of the \textit{direct path signal}, drawing inspiration from AIM~\cite{mao2018aim}, we employ separate speakers and microphones to capture the direct transmission in a controlled, quiet environment. This allows us to isolate and subsequently subtract the direct path signal from the received samples. The subtraction is executed by minimizing $||S-cS_d||$, where $S$ represents the samples received, $S_d$ denotes the direct signals pre-recorded, and $c$ is a scaling coefficient for optimal signal cancellation, which we have set at 0.9 in our experiments. This process effectively eliminates the \textit{direct path signal} between the speaker and microphone. Fig.~\ref{fig:filter_signal} illustrates this adjustment, showing the original raw signal (in blue) and the signal post-processing (in orange), where both background noise and direct path signal interference have been filtered out.

\begin{figure}[t]
\centering
\includegraphics[width=0.9\linewidth]{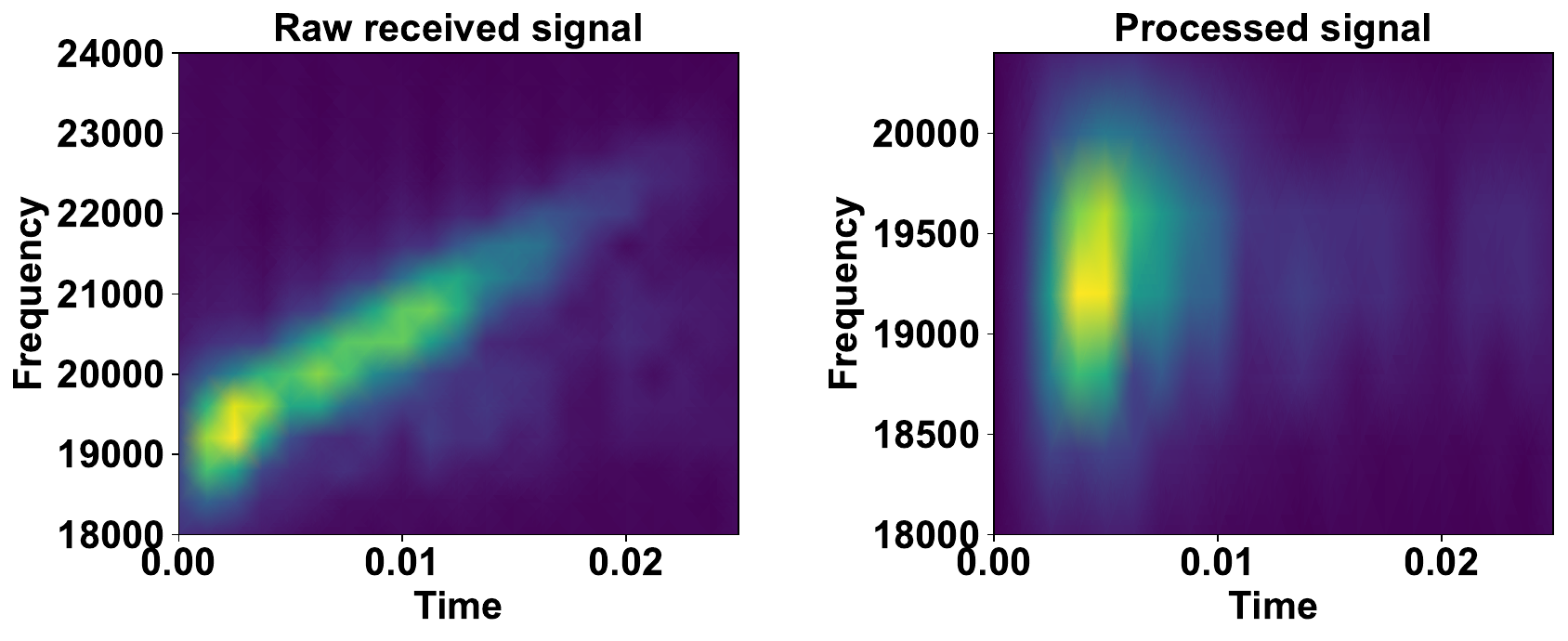}
\caption{The raw signal and the signal after noise removal.}
\label{fig:filter_signal_spectrogram}
\vspace{-10pt}
\end{figure}

Then, we consider eliminating the \textit{noisy echo signals} by identifying frequency shifts in received signals. The FMCW-based method is an essential technique for measuring distances and distinguishing between multiple echo sources based on the frequency shifts of the returned signals. In the scenario of a user interacting with a phone, it is reasonable to assume a relatively fixed and static distance between the phone and the user's face. This enables us to filter out \textit{noisy echo signals} originating from nearby obstacles at varying distances. Given that a comfortable viewing distance between a person's face and a smartphone typically ranges from 25 to 50 cm~\cite{zhou2018echoprint}, we utilize the equation in Eq.~(\ref{eq:fmcw}) to calculate the desired frequency shift as shown in Eq.~(\ref{eq:d_f}). This calculated shift helps us accurately identify and isolate the echoes emanating from the user's facial expressions while filtering out those from other objects.
\begin{equation}
    |\Delta f|=\frac{2RB}{Tv_0}.
    \label{eq:d_f}
\end{equation}
Thus, $|\Delta f|$ is between 235 Hz and 470 Hz. We further analyze the FMCW distance measurement resolution. Given the minimum measurable frequency shift $\Delta f_{min} = 1/T$, we can compute the resolution $d_r$ that FMCW separates mixed echoes as:
\begin{equation}
    d_r=\frac{v_0 \Delta f_{min} T}{2B}=\frac{v_0}{2B}.
    \label{eq:d_r}
\end{equation}
Thus, $d_r$ is $\frac{340m/s}{2\times 4000s^{-1}}=4.25~cm$. The resolution of the \textit{major echo signal} corresponding to a single sample is $\frac{v_0}{2F_s}=3.54~mm$, where $F_s$ is the sampling frequency 48 kHz. We employ the Short-Time Fourier Transform (STFT) using the Hann window to analyze the signal, which provides the complex amplitude of each frequency component. Considering the $|\Delta f|$ is within 500 Hz, we set the Butterworth filter with the critical frequencies between 190000 and 195000. Note that we only aim to mitigate the negative impact of multipath signals reflected from other objects, so we keep a tolerant frequency window size of 500 Hz. One advantage is that the relatively noisy input can help train a robust learning-based expression recognition model to adapt to various environments. Meanwhile, we acknowledge that with the increased distance between the face and the smartphone, according to the Eq.~\ref{eq:d_f}, $|\Delta f|$ should also increase. Fig.~\ref{fig:filter_signal_spectrogram} shows the spectrograms for the segmented major echo signal. In the left figure of Fig.~\ref{fig:filter_signal_spectrogram}, we show the raw received signal in the frequency domain, and the processed signal with frequency segmentation in the right figure. By computing the absolute values of the STFT output, we generate a spectrogram serving as the input in the learning-based model. 

\begin{figure*}[t]
\centering
\includegraphics[width=\linewidth]{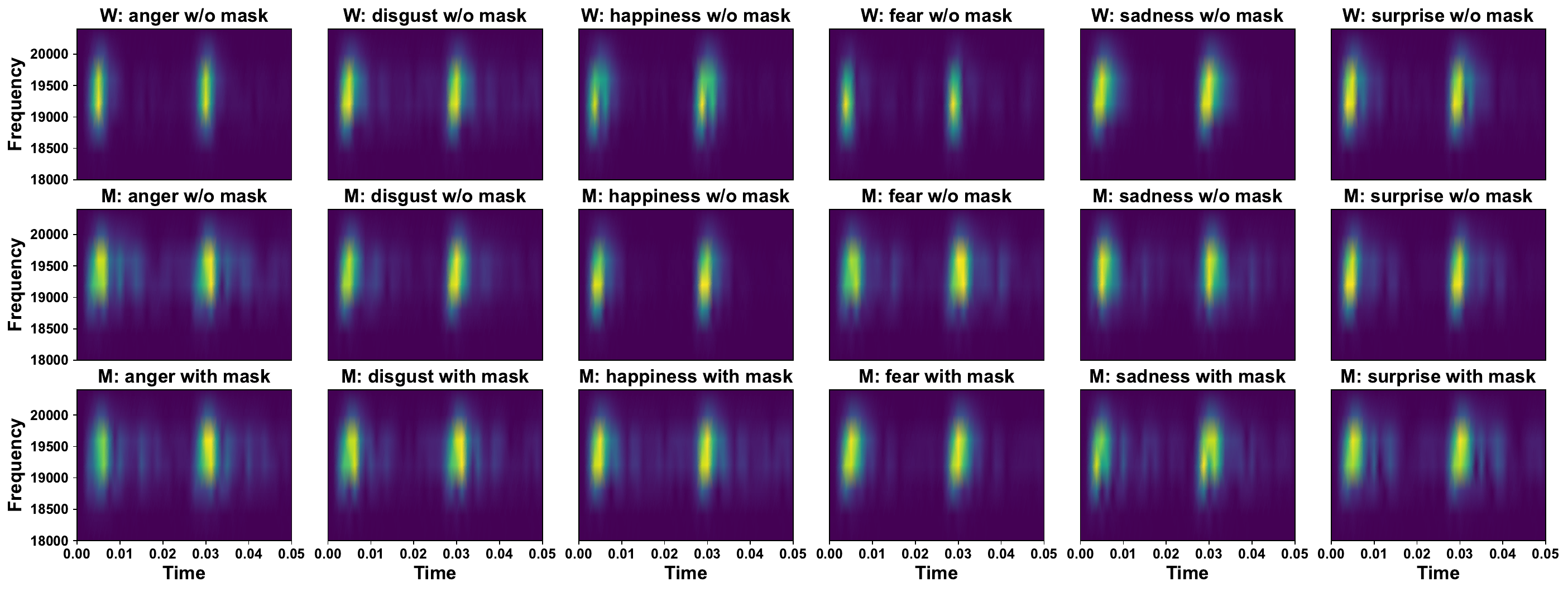}
\caption{The spectrogram of six expressions in 50 milliseconds. The first row is from a woman without a mask, the second row is from a man without a mask, and the third row is from the same man with a mask.}
\label{fig:spectrogram}
\vspace{-10pt}
\end{figure*}

\subsection{Domain Gap for Expression Recognition} 
Considering the variety of environmental noise and distinct facial expressions for different users, there are differences between the training and test datasets for expression inference, referred to as domain gap. Formally, a domain $\mathcal{D}=\{\mathcal{X}, P(X)\}$ includes the feature space $\mathcal{X}$ and marginal probability distribution $P(X)$. If two domains are different, they have different $\mathcal{X}$ or $P(X)$, but the label space is the same. We analyze the acoustic signal differences between different domains caused by environmental interference and user expression differences. Fig.~\ref{fig:spectrogram} shows the spectrograms for the segmented major echo signals associated with different facial expressions from two volunteers. From these visual representations, it is evident that different expressions produce distinct spectrograms for the same individual, highlighting the potential for recognizing and differentiating facial expressions based on acoustic signals.

First, various noises could enlarge the domain gap. It is challenging to completely eliminate noisy echo signals from various obstacles at different distances in certain scenarios. As mentioned before, in the design of \S, we establish the desired frequency shift, $|\Delta f|$, at 500 Hz. Nonetheless, minor multipath variations caused by body movements or objects interposed between the face and the phone are difficult to filter out. For instance, the second row in Fig.~\ref{fig:spectrogram} features spectrograms from an unmasked man, whereas the third row showcases the same individual wearing a mask. Subtle distinctions can be observed between the corresponding spectrograms for expressions such as ``surprise" across these two scenarios. This variation underscores the complexity of acoustic signal interpretation under different conditions. This is due to the resolution, $d_r$, of the FMCW is 4.25 cm as in Eq.~(\ref{eq:d_r}), which limits our ability to distinctly separate mixed echoes from face and surrounding objects such as facial masks. As a result, there are differences among acoustic signals from different domains. This factor could potentially affect the accuracy of capturing and analyzing the desired echo signals from facial expressions. 

Second, different users have distinct ways of showing their expressions, which widens the domain gap. For instance, as illustrated in the first and second rows of Fig.~\ref{fig:spectrogram}, individuals exhibit variations in how they express the same facial emotion, which is reflected in the differences in their respective spectrograms. Such variations lead to shifts in data distribution. The learning-based models usually have poor generalization ability when there is a new expression pattern for inference, which is caused by a distribution shift between the training and testing datasets~\cite{zhou2020xhar}.  Hence, it is crucial to align the distributions of the training and testing datasets, so as to reduce the domain gap and ensure the model can accurately generalize across different users' expressions.

\subsection{Acoustic Sensing Data Augmentation}
It is challenging to extract robust features that can accurately identify expressions from faint signals with limited acoustic samples. Collecting a diverse dataset from various populations helps learning-based models to robustly extract facial expression features. However, facial expression data collection and labeling are time-consuming activities, demanding considerable manual effort and coordination. 

\begin{algorithm}[t!]
\SetAlgoLined
\SetKwInput{KwInput}{Input}
\SetKwInput{KwOutput}{Output}
\caption{Acoustic Signal Data Augmentation}
\label{alg:aug}
\label{alg:augmentation}
\KwInput{initial dataset $D_{init}$, intra-person dataset $D_{intra}$, number of neighbors $K$, simulated distance $Dis$, classes $C$, weight $0\leq w \leq 1$}
\KwOutput{augmented dataset $D_{aug}$}
\ForEach{ $D_i \in$ $D_{init}$}{
    \If{IntraAug == True}{
        \For{$d=1$ to $Dis$}{
            $D_i^{d} = D_i*\frac{1}{\sqrt{d}}$
        }
        \ForEach{$D_i \in D_{intra}$}{
            $D_{aug} = w*D_i + \frac{1-w}{Dis}*\sum_{d=1}^{Dis}D_i^{d}$ 
        }
    }
    \Else{
        \ForEach{$D_i \in D_{intra}$}{
            $D_{inter} = D_{init}\backslash D_{intra} $ \\
            \ForAll{$D_{inter}\in C_{D_i}$}{
                $D_K^{dtw} = DTW_{N'bor}(D_i, D_{inter}, K)$ \\
            }
            $D_{aug} = w*D_i + \frac{1-w}{K/2}*\sum_{k=1}^{K/2}D^{dtw}_k$
        }
    }

\Return $D_{aug}$
}

\end{algorithm}
 
To address the above challenge, we design a novel data augmentation algorithm, which can be used to solve the issue of insufficient data by introducing altered versions of existing data and synthesizing new data based on the current dataset. Specifically, to expand the existing collected dataset $D_{init}$, we propose two augmentation strategies: intra-person and inter-person data augmentation (line 2 in Algorithm~\ref{alg:aug}). This expansion aids in improving the model's capacity to adapt to various acoustic settings and facial expressions. 

For intra-person data augmentation, following the inverse square law of sound propagation, we first transform the segment of the acoustic expression signal by a consistent length (lines 3-5 in Algorithm~\ref{alg:aug}). The signal's amplitude is adjusted by a factor proportional to the inverse square of the distance $Dis$ (e.g., 0.3 meters). Then, we create multiple versions of the acoustic signal $D_{aug}$ by summing the weighted average of raw signal $D_i$ with weight $w$ and transformed signals $D_i^d$ with weight $\frac{1-w}{Dis}$ (line 7 in Algorithm~\ref{alg:aug}). Considering our scenario when a user is holding a smartphone, a small device rotation at a fixed position creates negligible changes in the signal due to the omnidirectional nature of speakers and microphones, therefore, we only consider the changes in the device positions for acoustic signal augmentation.

For inter-person data augmentation, we first find the $K$-nearest neighbors of the target augmenting data $D_i$ in the same class $C_{D_i}$ as $D_i$. The $D_{inter}$ is the dataset from other persons than the person of $D_{intra}$. The implementation of $DTW_{N'bor}(\cdot)$ (line 14 in Algorithm) is similar to the K-Nearest Algorithm, but our goal is to find $K$ (\eg, $K=4$) samples from other persons for data synthesis. Specifically, we propose to use the dynamic time warping (DTW) distance to measure the similarity between different signals considering the inconsistent length of signals. From the $K$ samples, we randomly choose $K/2$ from the $K$ neighbors and allocate a weight of $\frac{1-w}{K/2}$ to each. To ensure the total weight sum is normalized, the original data sample $D_{i}$ receives the remaining weight $w$. Finally, we can get the augmented data $D_{aug}$ and enrich our acoustic facial expression dataset for contrastive external attention learning. 

\begin{figure*}[t]
\centering
\includegraphics[width=0.97\linewidth]{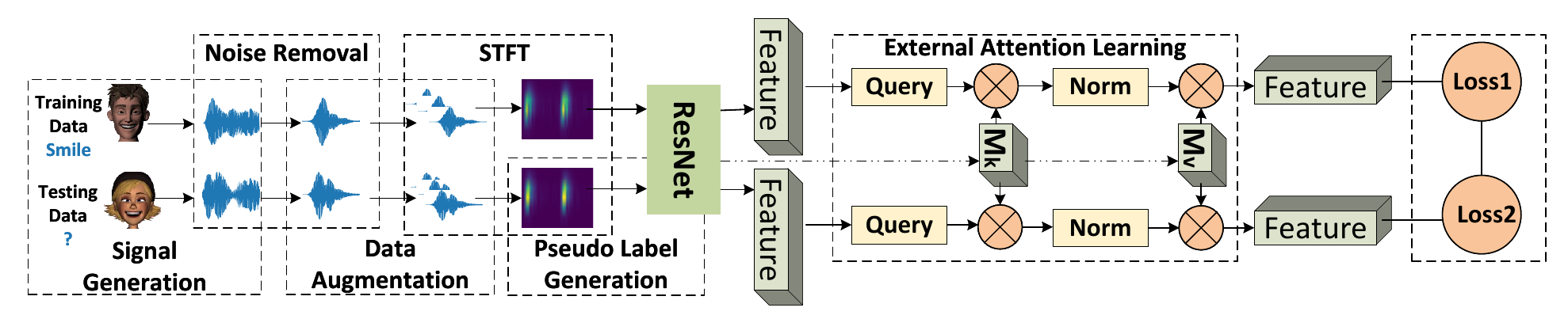}
\caption{The contrastive attention-based domain adaptation model for expression recognition, the losses are defined in Eq.~\ref{eq:loss}.}
\label{fig:model}
\vspace{-10pt}
\end{figure*}

\subsection{Contrastive Attention-based Domain Adaptation}
Image-based expression recognition models~\cite{cao2013facewarehouse, patil2016real, shen2016facial} excel at managing various backgrounds within images for expression feature extraction. Similarly, a learning-based model needs to extract expression-related features and separate acoustic nuances from external disturbances such as face masks in acoustic signals. Therefore, we design an acoustic feature extraction model based on contrastive learning, as shown in Fig.~\ref{fig:model} to identify and analyze facial expressions. The basic idea of contrastive learning is to learn distinctive representations by clustering positive pairs closer and distancing negative pairs. Here, the positive pairs refer to samples that have the same labels, and negative pairs have different labels. However, the samples from new test users lack labels, making it challenging to align samples in the latent space with contrastive learning. Therefore, we design a novel method for pseudo-label generation and cross-domain expression feature learning. 

\subsubsection{Pseudo Label Generation}
Consider a scenario where we have a training dataset from the source domain that includes fully labeled expression data, denoted as $D_s$. Meanwhile, we have a dataset for inference from the target domain that includes unlabeled data, represented by $D_t$, which shares the same categories as the source domain. For example, the $D_t$ can be obtained by collecting acoustic signals from a new user. Considering the differences between users and environments, there are distribution shifts between the source and target domains, degrading model inference performance. Therefore, we propose to align different domains by minimizing the distances between expression feature representations in latent space across different domains.

The initial challenge arises in forming positive pairs from the same categories within $D_s$ and $D_t$, particularly when the labels in $D_t$ remain unidentified. Drawing inspiration from DeepCluster~\cite{caron2018deep}, we generate pseudo labels for the unlabeled data in $D_t$ based on the highest category probability. Specifically, K-means clustering is presented to create these pseudo labels, then iteratively re-training the current model and refining the pseudo labels to reduce noise. We first pre-train the feature extraction model $g$ to make it converage on the source domain dataset. Then, we start by determining the centroid for each class in the source domain. As the labels are known in the source domain, we can compute the centroid embedding of each category as follows:
\begin{equation}
    c_k = \frac{1}{|D_s^k|}\sum_{x\in D_s^k}g(x),
    \label{eq:centroid}
\end{equation}
where $c_k$ is the centroid for class $k$, $|D_s^k|$ is the size of the $k$-th cateogory of data in the source dataset $D_s$. The function $g(x)$ produces the representation of an acoustic expression sample $x$ with the training function $g(\cdot)$. The clustering centroids can characterize the distribution of classes within the source domain. Pseudo-labels of the target domain sample $x_t$ are then assigned based on the nearest centroid as follows:
\begin{equation}
y_t = \argmin_k ||g(x_t), c_k||_2,
\label{eq:tar_label}
\end{equation}
We calculate the $l_2$ distance between the representation embedding $g(x_t)$ and the centroid embedding $c_k$. Consequently, this allows for the generation of pseudo labels for the target domain dataset. Meanwhile, we acknowledge the misassigned pseudo-labels can negatively slow down the model converge process. Therefore, we propose to apply the median absolute deviation (MAD)~\cite{yang2021cade, leys2013detecting} to measure the variability of test samples. Specifically, we first get the latent feature representations of training and test samples. Then, we can calculate the average embedding in each category $k$ of the training samples as $\bar{Z_k} = \frac{1}{n}\sum^n_i Z_k^i$, where $i$ is the index of training samples in each category. The distance between each sample embedding and its category average embedding is $d_k^i=|Z_k^i-\bar{Z_k}|$. The median of the distance is computed as $\hat{d_k} = median(d_k^i)$, and the MAD is computed as $MAD_k = median(|d_k^i-\hat{d_i}|)$. For the test samples, we can get the distance between each test sample embedding and the training category average embedding as $d_j = |Z_j-\bar{Z_k}|$, where $j$ is the index of testing samples. Finally, the drifting score for each test sample is computer as $T_j^k = \frac{d_j - \hat{d_k}}{MAD_k}$. If $T_j^k$ is larger than a threshold of 3.5, we will filter out the test samples and their assigned pseudo-labels. Finally, the formation of positive and negative pairs uses datasets from both the source and target domains. The model parameters are iteratively updated by minimizing the loss function outlined in Eq.~(\ref{eq:loss}) below. With the increase in training iterations, the model $g(\cdot)$ gains better feature learning ability, so as to better depict the categories of samples in the target domain.

\subsubsection{Attention-based Expression Learning}
As previously noted, completely eliminating all noisy echoes is nearly unfeasible. Therefore, the model needs to gain the ability to differentiate the characteristics of facial expression echoes from those of the background noise. During each expression (lasting approximately 1 second), multiple data samples (0.1 seconds each) are produced. Identifying correlations among these samples could help the model concentrate on the consistent and common features of acoustic facial expressions. However, while self-attention is commonly employed to learn robust features, it exhibits quadratic complexity and fails to account for possible correlations between different samples. 

To overcome the mentioned limitation, we propose to employ external attention~\cite{guo2021beyond} to depict key features and implicitly learn correlations across all expression samples. Following the notation in Eq.~(\ref{eq:attention}), we initially calculate the attention map $A = QM_k^T$ by multiplying the query vector $Q$ with the externally learnable, transposed key matrix $M_k \in \mathcal{R}^{S\times d}$. Here, $Q$ is derived from the projection of a feature map $F \in \mathcal{R}^{N\times d}$, where $N$ represents the number of feature elements, and $S$ and $d$ are hyper-parameters. We apply normalization to the attention map $A$, which is then multiplied by another external value matrix $M_v$. Both $M_k$ and $M_v$ are produced by additional linear layers, which can be fine-tuned through back-propagation during training across the full dataset. The resulting attention-enhanced feature map, $F_{out}$, is as follows:
\begin{equation}
    F_{out} = Norm(QM_k^T)M_v.
    \label{eq:ex_att}
\end{equation}
Ultimately, we achieve a refined feature map with linear complexity, represented as $O(d \cdot S \cdot N)$, making it well-suited for resource-limited mobile devices. 

\begin{algorithm}[t!]
\SetAlgoLined
\SetKwInput{KwInput}{Input}
\SetKwInput{KwOutput}{Output}
\caption{Contrastive Attention-based Cross Domain Acoustic Expression Representation Learning}
\label{alg:rep}
\KwInput{source dataset $D_s$, target dataset $D_t$, epoch $E$, iterations $K$ per epoch, weight $\lambda$, contrastive attention-based model $g$}
\KwOutput{source and target representations $Z^s$ and $Z^t$}
\For{$e=1$ to $E$}{
    Calculate centroids in target domain using Eq.~\ref{eq:centroid}\\
    Update pseudo labels for target data using Eq.~\ref{eq:tar_label}\\
    \For{$k=1$ to $K$}{
        \For{each batch}{
            Extract features with $f$ based on external attention in Eq.~\ref{eq:ex_att}\\
            Compute $L_{ce}$ for each batch from $D_s$\\
            Compute $L_{con}$ from $D_s$ and $D_t$ using Eq.~\ref{eq:domain} \\
            Compute $\lambda\mathcal{L}_{ce}(\theta; D_s, D_t) + (1-\lambda) \mathcal{L}^t_\text{con}(\theta; D_s, D_t)$         
        }
        Back-propagate and update $\theta$ of model $g$
    }
}
\For{each batch $X_{batch}$}{
    Generate source domain expression representation $Z_{batch}^s = g(X_{batch}^s)$ for $D_s$\\
    Generate target domain expression representation $Z_{batch}^t = g(X_{batch}^t)$ for $D_t$
} 
\Return $Z^s$ and $Z^t$
\end{algorithm}

\subsubsection{Feature Alignment for Domain Adaptation}
It is reasonable to assume that samples belonging to the same class cluster closer together in the latent space, whereas samples from different classes are more distant, regardless of their domain of origin. Utilizing the augmented dataset, pseudo labels, and the attention-based learning model, we implement contrastive learning to reduce domain discrepancies by aligning facial expression features between the training and testing datasets.

Specifically, when presented with an acoustic facial expression sample $x_s$ from the source domain and a sample $x_t$ from the target domain, we aim to minimize the distance between $x_s$ and $x_t$ if they belong to the same class, while maximizing the distance between samples from different classes. This process yields domain-independent expression representations. In line with the supervised contrastive loss outlined in Eq.~(\ref{eq:contrastive2}), we define the domain adaptation contrastive loss as follows:
\begin{equation}
\begin{split}
        \mathcal{L}^t_\text{con} = \sum_{i\in I_t}\frac{-1}{|P_s(y_t^i)|}\sum_{p\in P_s(y_t^i)}log\frac{exp(z_t^i\cdot z_s^p/\tau)}{\sum_{a\in I_s} exp(z_t^i\cdot z_s^a/\tau)},
\end{split}
\label{eq:domain}
\end{equation}
where $I_t$ represents the set of target samples within a batch, $I_s$ denotes the set of source samples, and $P_s(y_t^i)$ refers to the indices of all positive samples from the source domain. A positive sample is defined as having a label that matches the pseudo label of the target anchor sample $x_t$. The domain adaptation contrastive loss is designed to align the expression representations from the target domain with those from the source domain. Finally, we formulate the loss function for learning acoustic expression representations as follows:
\begin{equation}
    \argmin_\theta \lambda\mathcal{L}_{ce}(\theta; D_s, D_t) + (1-\lambda) \mathcal{L}^t_\text{con}(\theta; D_s, D_t),
    \label{eq:loss}
\end{equation}
where $\mathcal{L}_{ce}$ denotes the cross-entropy loss applied to the dataset $D_s$, $\theta$ denotes the model parameters, and  $\lambda$ serves to balance the two loss terms. Here the $\lambda$ is defined as follows:
\begin{equation}
    \lambda(t) = \frac{exp(w_1(i-1))}{exp(w_1(i-1)) + exp(w_2(i-1))},
    \label{eq:lambda}
\end{equation}
where $w_1(i-1) = \frac{L_{ce}(i-1)}{L_{ce}(i-2)}$, $w_2(i-1) = \frac{L_{c}^t(i-1)}{L_c^t(i-2)}$. $i$ is the iteration index during the model training process. For $i=1$ and $2$, $w_1$ and $w_2$ are 1. $\lambda$ is computed by the softmax of loss increment ratio $w(i-1)$ in each learning objective.

In summary, the designed contrastive external attention-based model is outlined in Algorithm~\ref{alg:rep}. The process begins with data augmentation, followed by generating pseudo labels for target domain acoustic samples in each epoch (lines 2-3). Subsequently, we reduce the loss and perform back-propagation to update the model $f$ (lines 4-12). Upon completion of training, model $g$ aligns features for effective domain adaptation, thereby minimizing distribution shifts. The trained model $g$ is then used to produce acoustic expression representations $Z^s$ and $Z^t$ (lines 14-17). Finally, a classifier is trained on the source domain representations $Z^s$ to predict labels for the target domain representations $Z^t$. This approach significantly improves the performance of the expression recognition model across different users.

\section{Implementation}
\label{implementation}
In this section, we introduce the data collection process, the software, and the hardware setup for acoustic expression recognition implementation.

\subsection{Data Collection}
We collect acoustic facial expression data from 20 volunteers (16 males and 4 females) in two main time periods (May 2022 and May 2024). To ensure a diverse range of facial expressions, these volunteers vary in skin color from various regions of the world, including Asia, North America, and Europe, with ages spanning from 20 to 38 years. During the data collection, participants were permitted to wear glasses, hats, and other accessories. To mimic a variety of real-life situations, we gathered data in diverse settings (e.g., offices, dining halls, gardens) featuring varying levels of background noise. For instance, data collection occurred in an office environment amid sounds of people conversing, participating in online meetings, and computer alarms beeping.

\begin{figure}[t]
\centering
\includegraphics[width=0.8\linewidth]{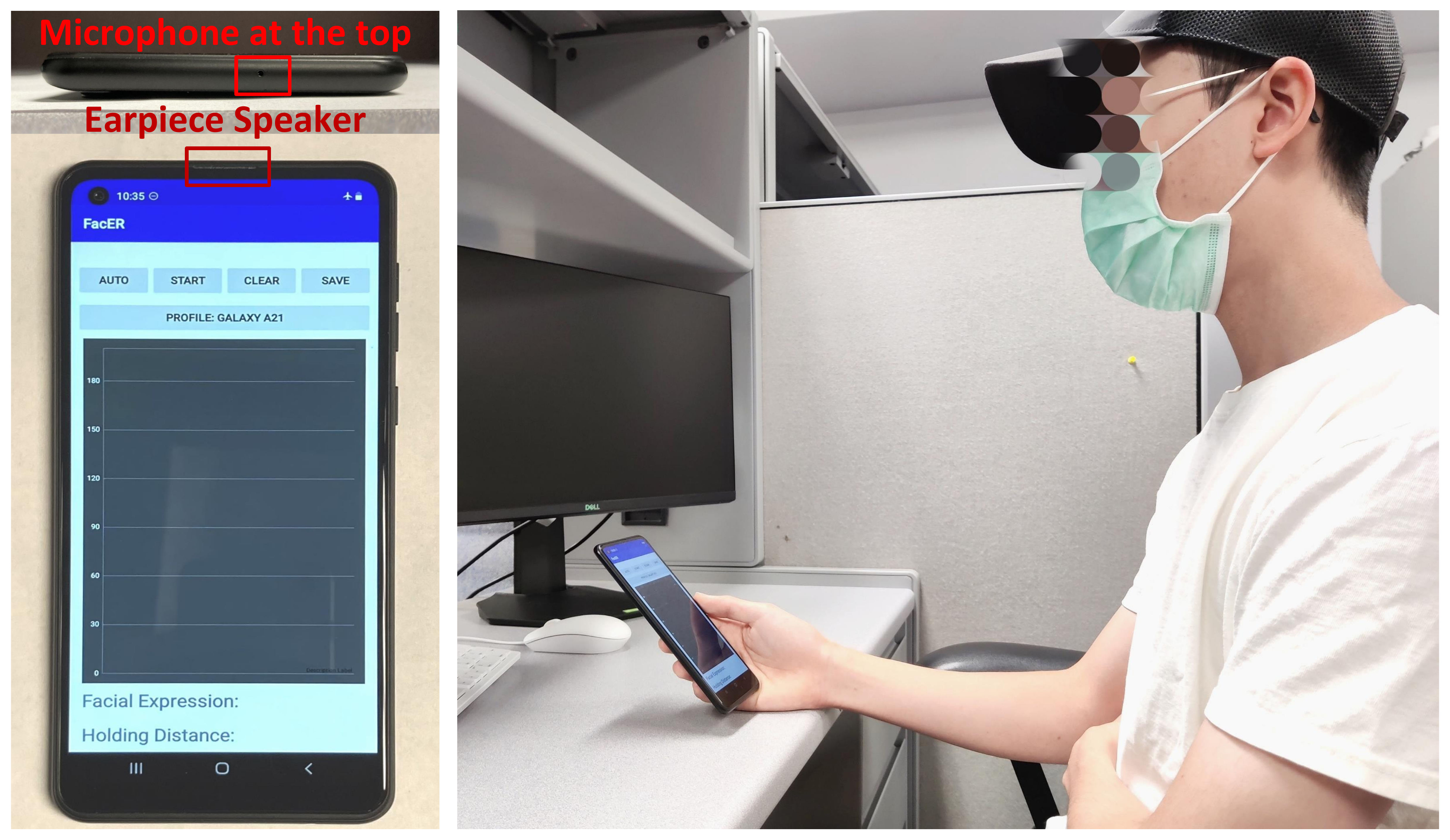}
\caption{A smartphone and a volunteer for data collection.}
\label{fig:setup}
\end{figure}

At the beginning of the data collection, we show the six standard facial expressions: anger, disgust, fear, happiness, sadness, and surprise. Each volunteer begins with a neutral expression and then proceeds to exhibit their unique style of these six expressions. Fig.~\ref{fig:setup} illustrates a volunteer and the smartphone setup used during the data collection. Volunteers are encouraged to hold smartphones in their most comfortable manner while looking at the smartphone screen. Given that face masks are commonly used in our daily lives, our study also accounts for expression recognition when participants are wearing masks. For each expression, we collect approximately 5 seconds of data with participants wearing masks and another 5 seconds without masks.

The data collection for each expression is repeated 10 times per individual, with breaks included throughout the sessions in the first data collection period. In the second data collection period, we only ask volunteers to repeat 2 times for each facial expression. An independent observer records the label for each acoustic expression sample to serve as ground truth. The entire data collection spans about a week. In total, we extracted 22,054 samples with a window size of 0.25 seconds. This window size was chosen to balance detail and clarity: too small a window might not capture the full dynamics of facial muscle movements, while too large a window could blend the fleeting variations between different expressions.

\subsection{Experimental Setup}
We utilized two Android smartphones, the Samsung Galaxy A21, and OnePlus 8T, for collecting acoustic signals. The sensing signal is a chirp signal modulated between 19-23 kHz, which is emitted from the earpiece speaker of the smartphone. The app for acoustic sensing data collection leverages frameworks from LibAS~\cite{tung2018cross} and Chaperone~\cite{chen2020chaperone}, which are designed for acoustic sensing applications. LibAS, in particular, streamlines various signal processing tasks, including synchronization, which identifies the starting point of transmitted signals within the received audio. 

We utilize the SciPy library for signal-processing tasks including the Butterworth filter and short-time Fourier transform. Our designed contrastive attention-based expression recognition model is built on the ResNet-18 architecture~\cite{he2016deep}. Specifically, we use the external attention-based ResNet-18 model to implement classification. We use the SGD with a momentum of 0.9 and the learning rate is 0.1. We implement Algorithm~\ref{alg:rep} to extract feature representations, and then we use a linear classifier with hidden layer size 256 to implement expression classification. The model is implemented using Pytorch. The training is conducted on an Ubuntu 20.04 Server, equipped with Intel(R) Xeon(R) Gold 5218R CPUs at 2.10GHz, and RTX A6000 GPUs.

\section{Evaluation}
\label{evaluation}
In this section, we evaluate the effects of various elements (\eg, location, time, people, mask) on the performance of \S in identifying different acoustic facial expressions.

\begin{figure}[t]
\centering
\begin{subfigure}{0.24\textwidth}
    \includegraphics[width=0.95\linewidth]{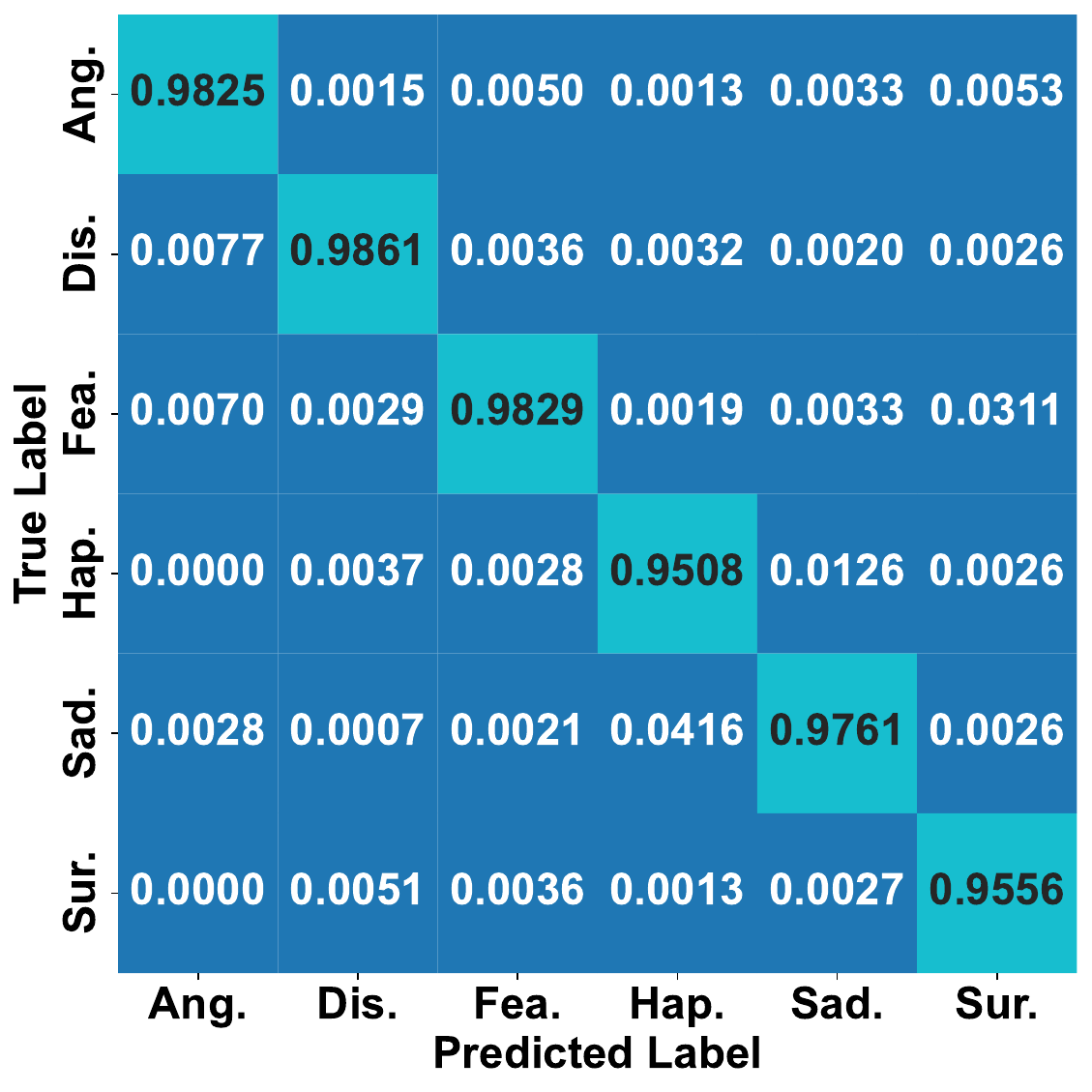}
    \caption{Dataset mix testing.}
    \label{fig:case1}
\end{subfigure}
\begin{subfigure}{0.24\textwidth}
  \includegraphics[width=0.95\linewidth]{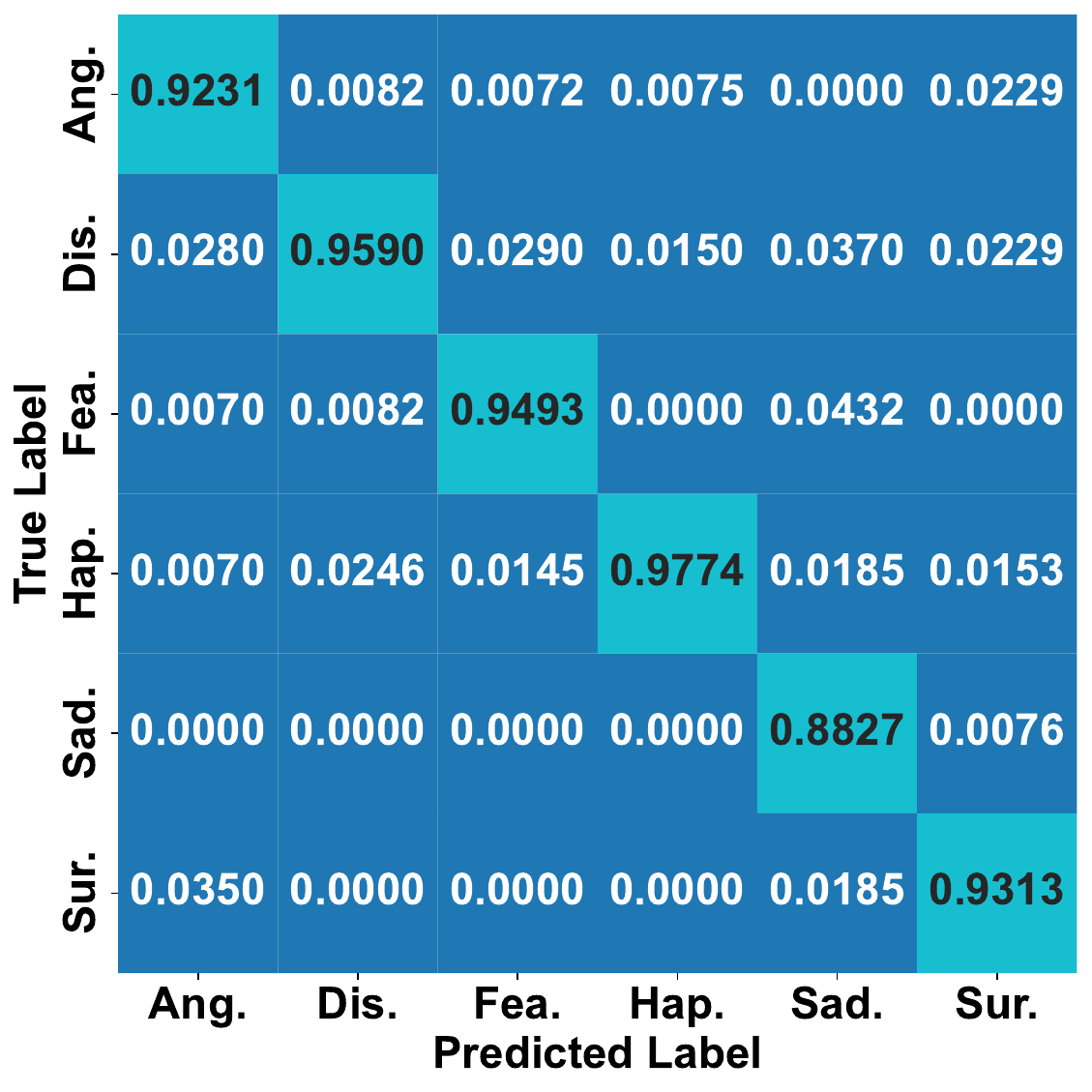}
  \caption{Leave-one-out testing.}
  \label{fig:case1-1}
\end{subfigure}
\caption {Evaluations of different cases. (a): training and testing on the mixed dataset; (b): training on data from 19 people and testing on the remaining one user; The tick labels are angry, disgust, fear, happiness, sadness, and surprise. The values in the heat map are normalized over the predicted (columns) conditions and rounded.}
\label{fig:heatmap_1}
\end{figure}

\subsection{User Dependent Evaluation}
  
\textbf{Case 1a.} We begin by examining a simple scenario in which we collect and label acoustic data from a group of users. Our objective is to recognize facial expressions from this specific group, referred to as \emph{dataset mix testing}. We allocate 80\% of the entire dataset for training and the remaining 20\% for testing. The performance of our model is shown with an accuracy heat map in Figure~\ref{fig:case1}. \S can identify each acoustic facial expression with over 95\% accuracy. Additionally, we performed 10-fold cross-validation, achieving an average testing accuracy of 97.2\% with a standard deviation of 1.39\%. The results are expected because the training and testing datasets share the same distribution, allowing the model to fit the data easily. Next, we conducted a leave-one-user-out test, where existing studies~\cite{wang2024uface, gao2021sonicface} struggle to achieve comparable performance as in mix testing. As shown in Figure~\ref{fig:case1-1}, the leave-one-user-out test resulted in an average accuracy of 93.7\%, which achieves comparable performance as the dataset mix testing. By comparison, UFace~\cite{wang2024uface} achieves 87.8\% average accuracy in mix testing while only achieving 61.65\% accuracy for new users without fine-tunning. The results showcase the effectiveness of \S for new user domain adaptation. An extensive analysis of user-independent evaluation cases is provided in Section~\ref{subsec:use-indep}.

\begin{figure}[t]
\centering
\includegraphics[width=\linewidth]{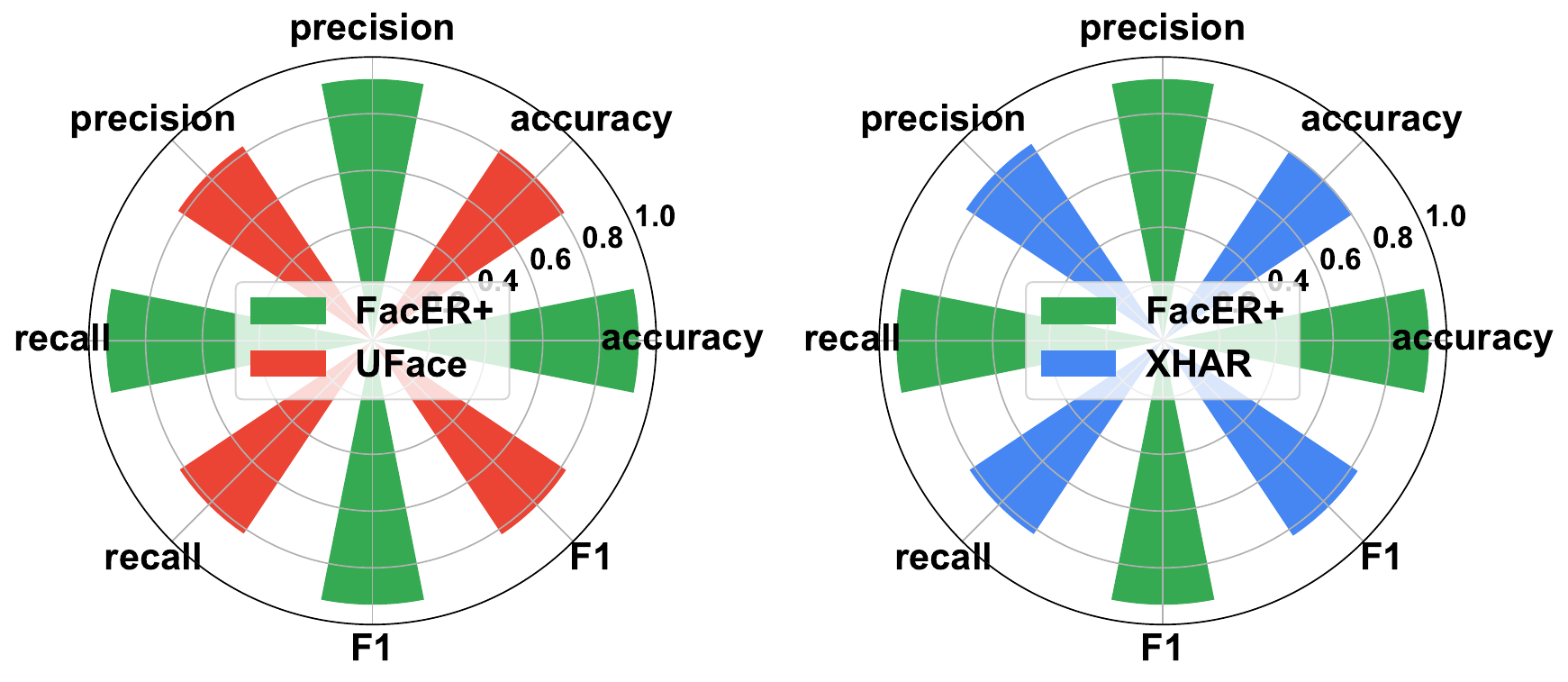}
\caption{Case 1b: The bars of location factor evaluation.}
\label{fig:polar}
\end{figure}

\textbf{Case 1b.} Next, we evaluate the impact of environmental factors on the performance of \S. We test the model using data collected from three different locations: an office, a dining hall, and a garden. We perform a \emph{leave-one-place-out} evaluation by training the model on data from two locations and testing it on data from the third location. We compare the performance of \S against two baselines: (i) DFNet in UFace~\cite{wang2024uface}, a multi-view CNN architecture for extracting feature representations, and (ii) XHAR~\cite{zhou2020xhar}, an adversarial training-based domain adaptation method for human activity recognition. Since adversarial training is a prominent approach for cross-domain adaptation, we selected XHAR as a baseline and adapted its methodology for facial expression recognition.

We present the average accuracy, precision, recall, and F1 scores across the three locations in Fig.~\ref{fig:polar}, which is depicted as a bar plot on a polar axis. The accuracy for UFNet, and XHAR are 81.3\% and 80.3\%, with standard deviations of 0.013, and 0.016, respectively. \S achieves an average accuracy of 93.8\%, slightly lower than the mix testing method (97.2\%). These results indicate that location induces distribution shifts and impacts model performance due to noise from various obstacles. Despite this, \S still outperforms existing methods. For example, \S achieves an F1 score of 93\%, outperforming the XHAR method (82.6\%), which is attributed to our designed contrastive attention-based domain adaptation algorithm. This demonstrates that \S can learn consistent and robust acoustic facial expression features even in noisy environments with diverse types of noise.

\begin{figure}[t]
\centering
\includegraphics[width=0.8\linewidth]{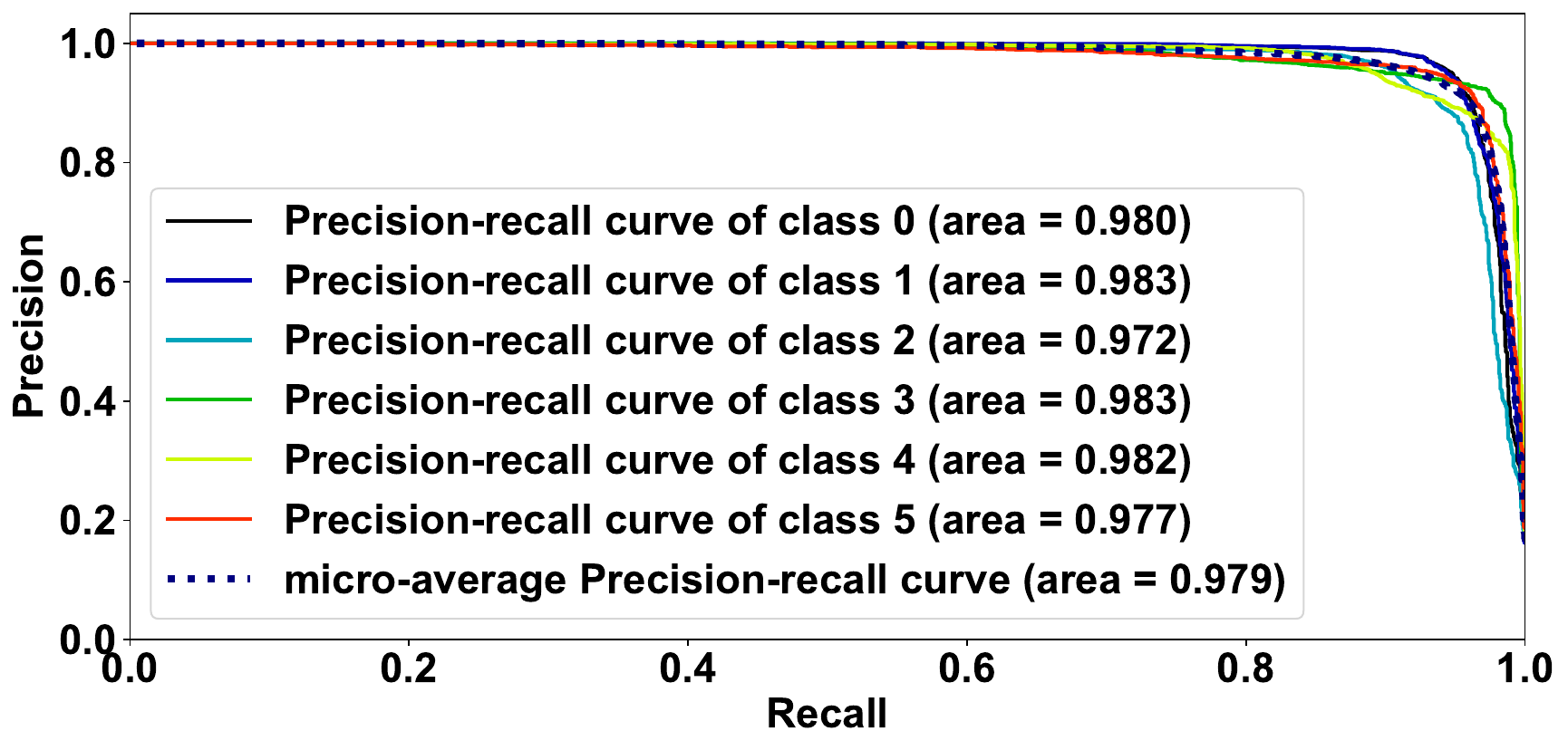}
\caption{Case 1c: The precision-recall of time factor evaluation.}
\label{fig:pr}
\end{figure}

\begin{figure}[t]
\centering
\includegraphics[width=0.8\linewidth]{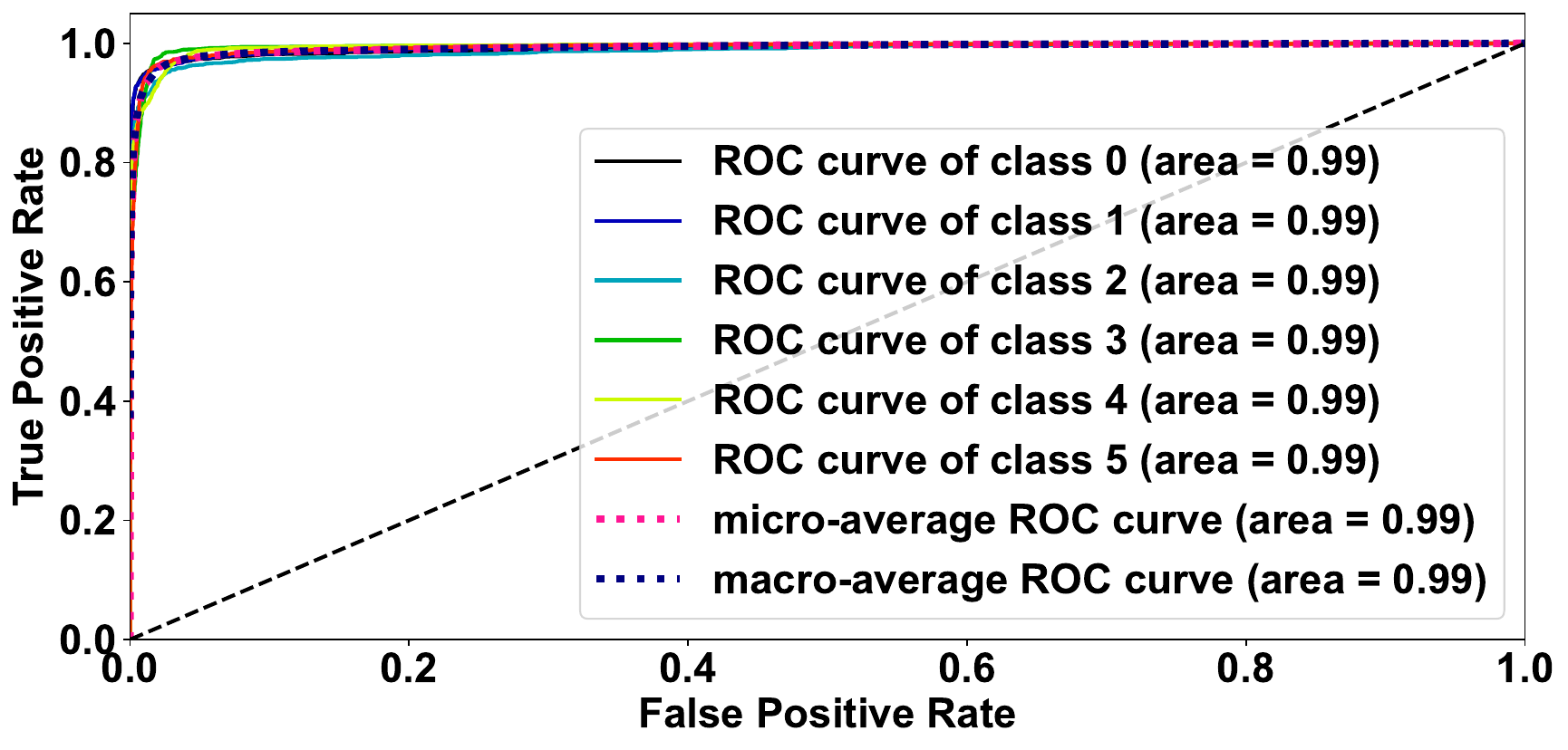}
\caption{Case 1c: The ROC curve of time factor evaluation.}
\label{fig:roc}
\end{figure}

\textbf{Case 1c.} We consider a more complex scenario involving time variation. The same facial expression might not be consistent over time; for example, a person might show a wide smile at one moment and a gentle smile at another to express happiness. To evaluate the impact of time on acoustic facial expression recognition, we collected new data in May 2024, following the initial training dataset collection two years earlier. Specifically, the training dataset was collected in May 2022, and the test dataset was collected in May 2024 from 6 volunteers, who were among the 20 volunteers from May 2022. It is important to note that the newly collected dataset is used solely for testing purposes. 

The evaluation results are shown in Fig.~\ref{fig:pr} and Fig.~\ref{fig:roc}. The precision-recall score curve in Fig.\ref{fig:pr} illustrates the tradeoff between precision and recall across various thresholds. The area under the curve, calculated using average precision (AP), is determined by the formula: $AP = \sum_n (R_n - R_{n-1})P_n$, where $P_n$ and $R_n$ represent the precision and recall at the $n_{th}$ threshold, respectively, as set automatically by Scikit-plot\cite{scikit}. Precision is defined as $\frac{tp}{tp+fp}$ and recall as $\frac{tp}{tp+fn}$, where $tp$ represents true positives, $fp$ false positives, and $fn$ false negatives. For different classes, \S achieves an AP score of at least 0.97. A high precision-recall area under the curve indicates both high recall and high precision, reflecting low false-positive and false-negative rates. The results show that \S still maintains high performance when recognizing expression after a long period, demonstrating the efficacy of the designed model in acoustic expression recognition.

In Fig.~\ref{fig:roc}, we illustrate the receiver operating characteristic (ROC) curve, showcasing the performance of \S across various classification thresholds. The thresholds are automatically set by~\cite{scikit}. Reducing the threshold leads to more items being identified as positive, which increases both the true positive and false positive rates. The area under the ROC curve (AUC) measures the entire two-dimensional space underneath the ROC curve, representing the likelihood that the model will correctly rank a randomly selected positive example over a negative one. An AUC of 0.0 indicates entirely incorrect predictions, while an AUC of 1.0 signifies completely accurate predictions. As depicted in Figure~\ref{fig:roc}, \S achieves a high AUC score of 0.99 for different expression classes. Therefore, \S still has a low false-positive rate when considering the impact of time factor in expression recognition.

\begin{figure*}[t]
\centering
\begin{subfigure}{0.245\textwidth}
  \includegraphics[width=0.95\linewidth]{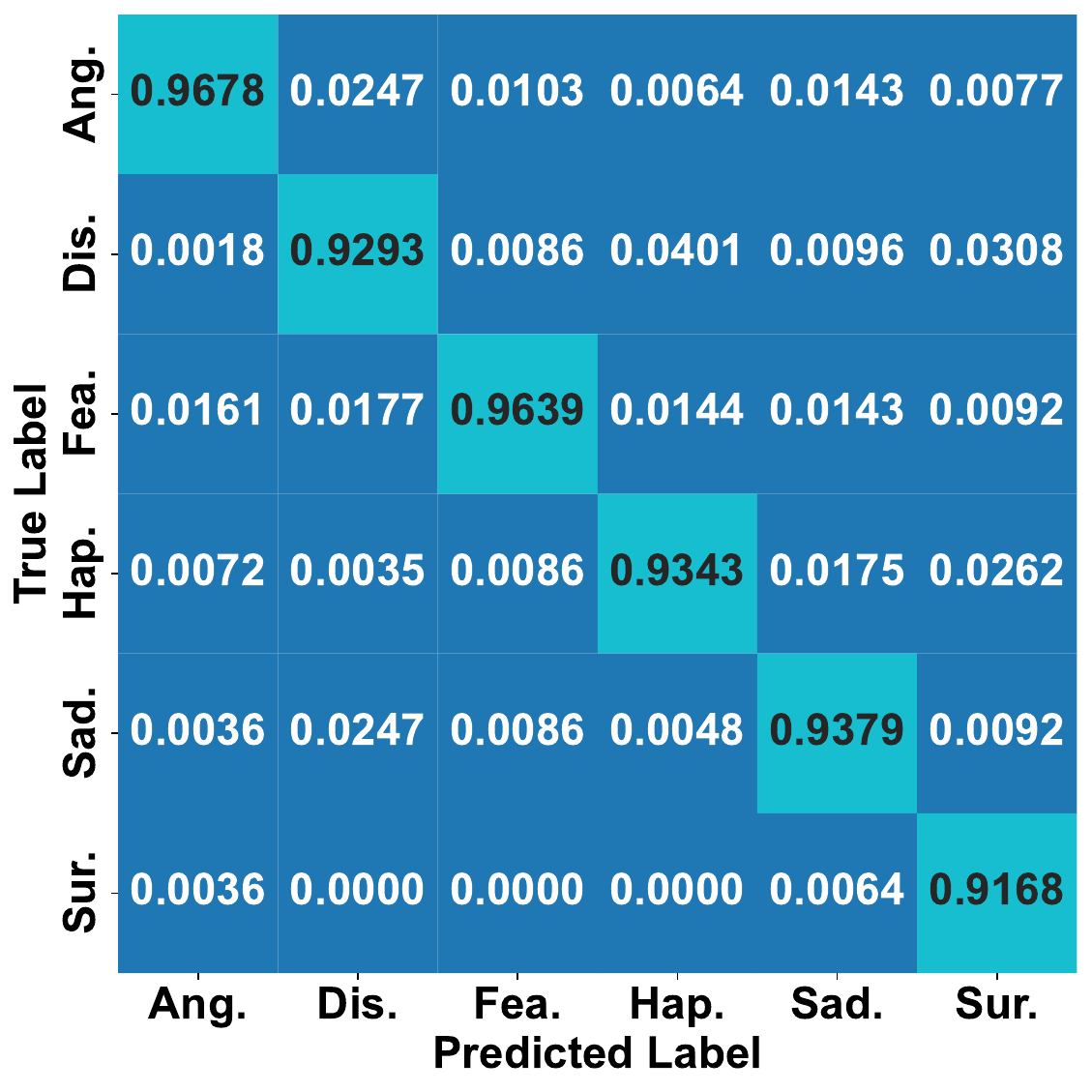}
  \caption{Case 2: Testing on Women's.}
  \label{fig:case2}
\end{subfigure}
\begin{subfigure}{0.245\textwidth}
    \includegraphics[width=0.95\linewidth]{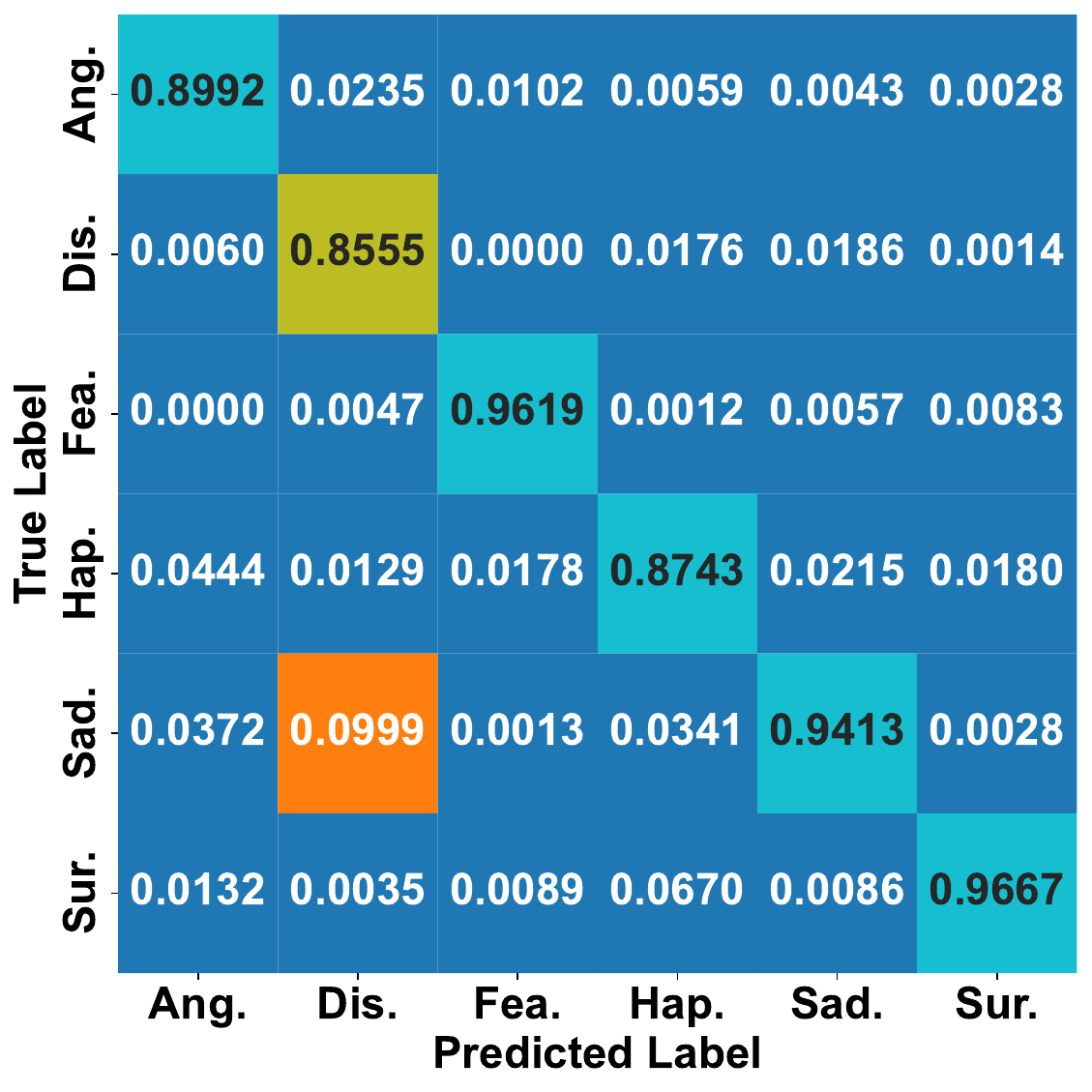}
    \caption{Case 3: Testing on 10 users.}
    \label{fig:case3}
\end{subfigure}
\begin{subfigure}{0.245\textwidth}
    \includegraphics[width=0.95\linewidth]{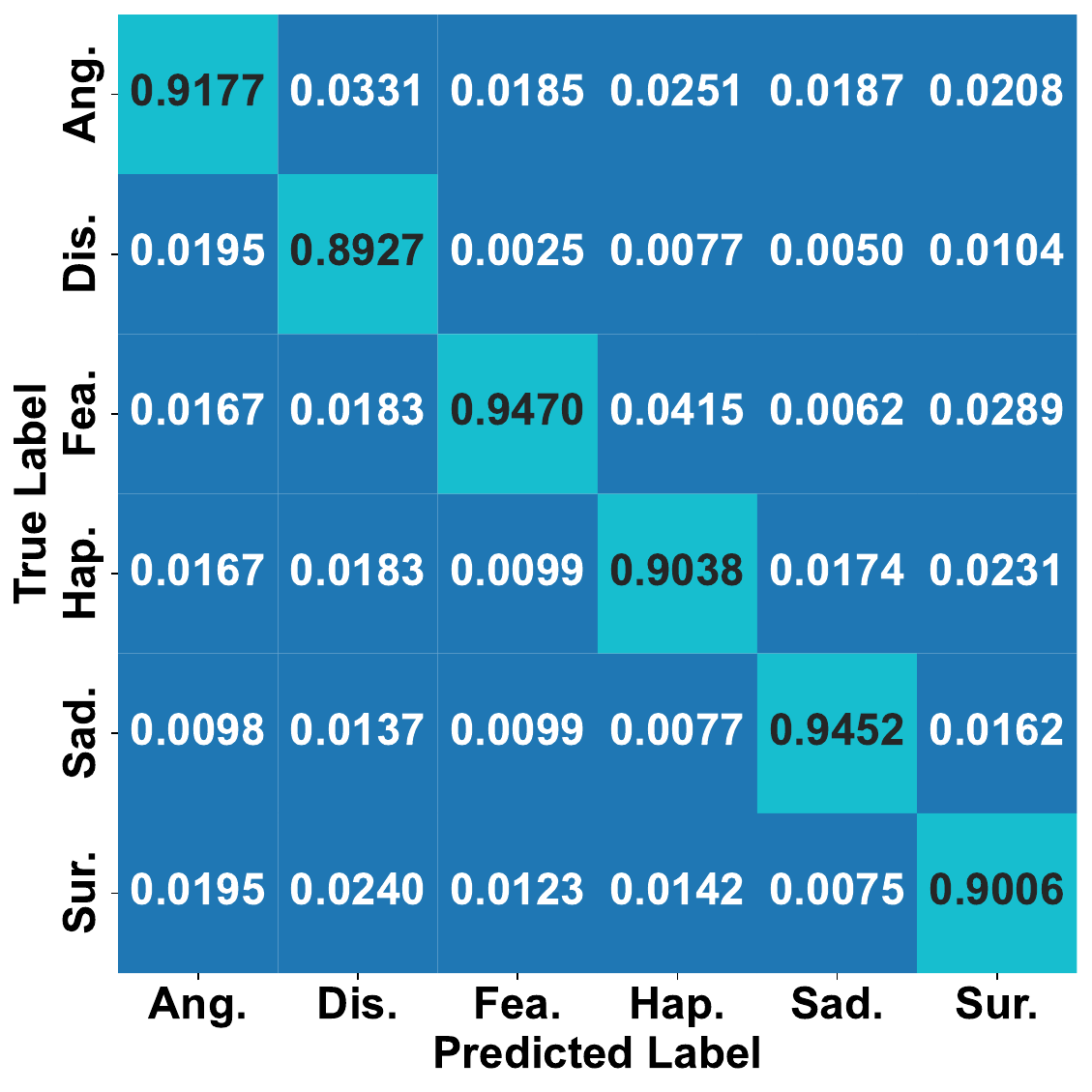}
    \caption{Case 4: Testing on Mask's.}
    \label{fig:case4}
\end{subfigure}
\begin{subfigure}{0.245\textwidth}
    \includegraphics[width=0.95\linewidth]{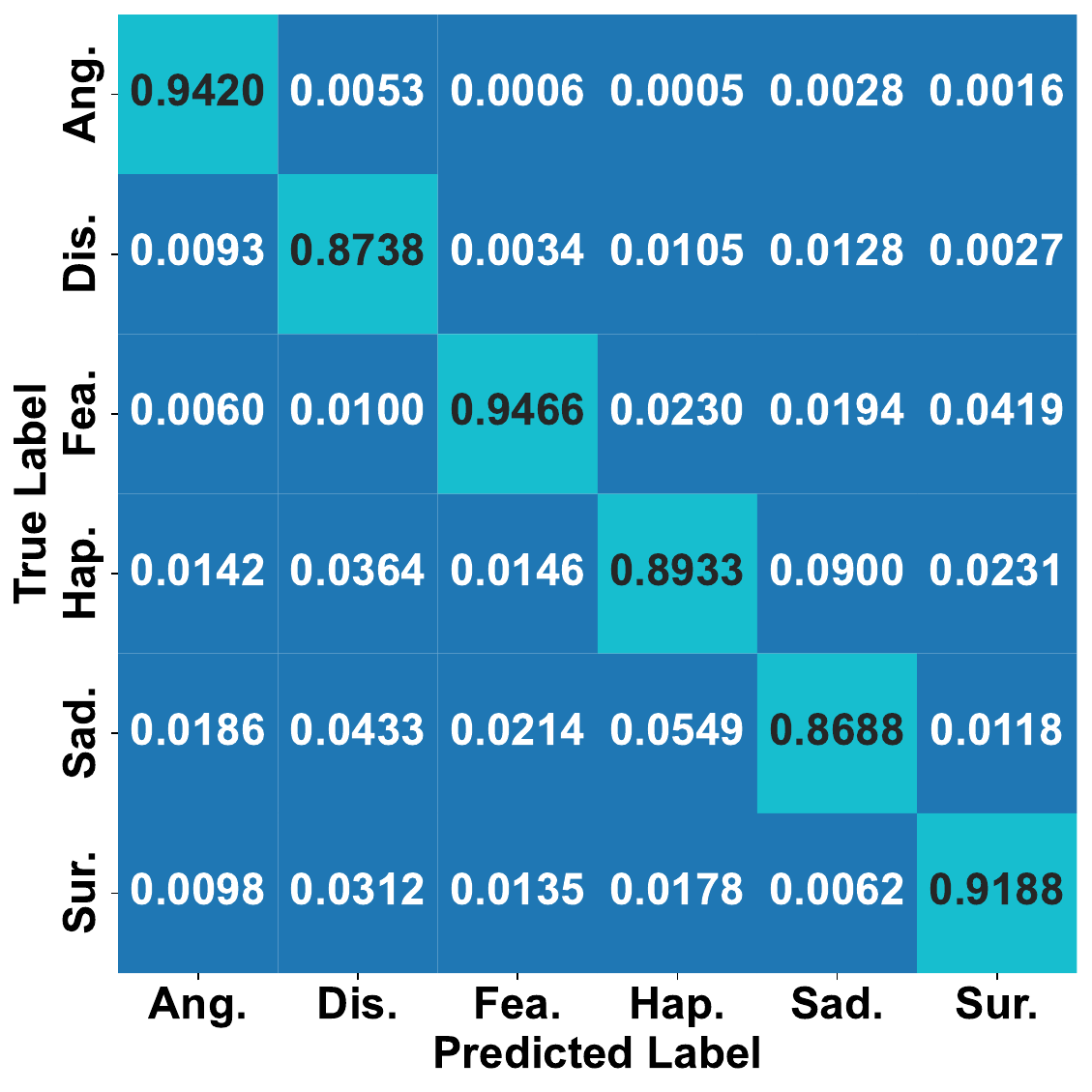}
    \caption{Case 5: Testing on No Mask's.}
    \label{fig:case5}
\end{subfigure}
\caption {Evaluations of different cases. Case 2: training on men's dataset and testing on women's dataset; Case 3: training on data from 10 users and testing on data from other 10 users. Case 4: training on users without masks and testing on users with masks; and Case 5: training on users with masks and testing on users without masks.  The tick labels are angry, disgust, fear, happiness, sadness, and surprise. The values in the heat map are rounded.}
\label{fig:heatmap_2}
\end{figure*}

\begin{figure}[t]
\centering
\begin{subfigure}{0.24\textwidth}
\centering
    \includegraphics[width=0.95\linewidth]{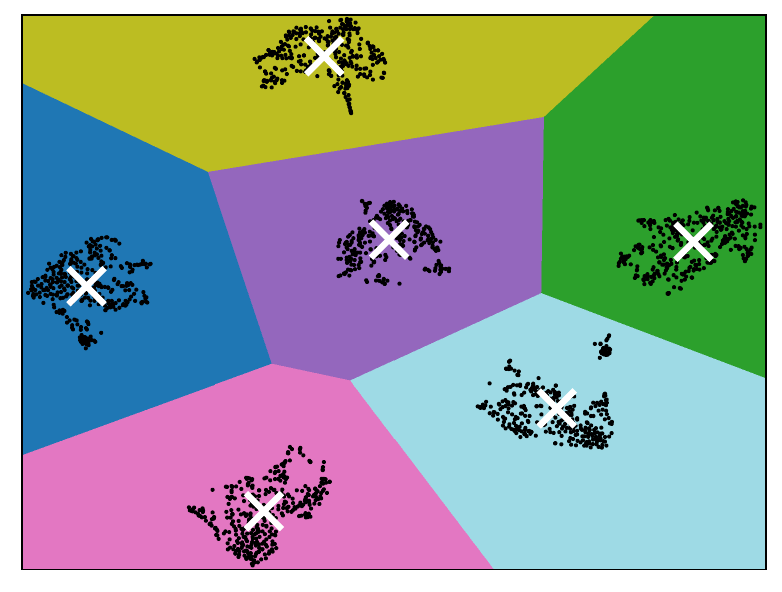}
    \caption{Clustering on the training set.}
    \label{fig:train}
\end{subfigure}
\begin{subfigure}{0.24\textwidth}
\centering
  \includegraphics[width=0.95\linewidth]{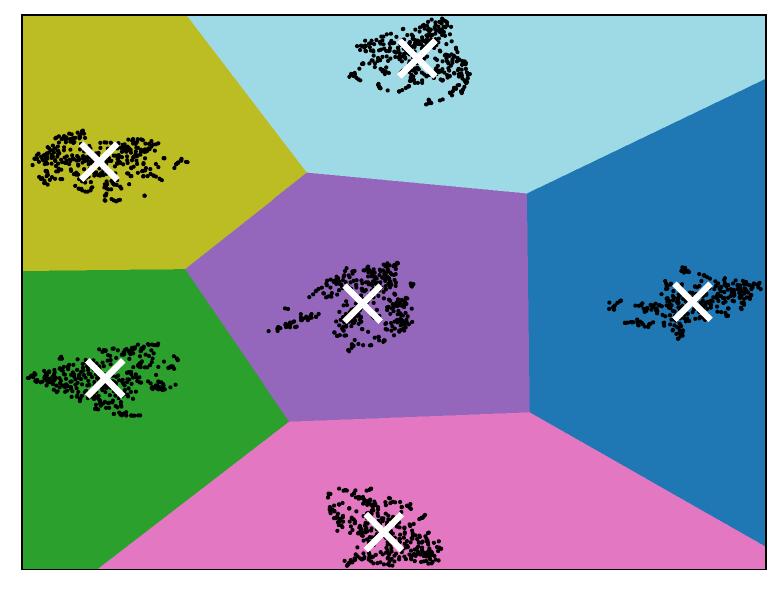}
  \caption{Clustering on the testing set.}
  \label{fig:test}
\end{subfigure}
\caption {K-means clustering on 2,048 sampled representations, which are processed with TSNE dimension reduction.}
\label{fig:contrastive_points}
\end{figure}

\subsection{User Independent Evaluation}
\label{subsec:use-indep}
\textbf{Case 2.} We now explore a more generalized scenario where the model is trained on one group of users and tested on another. Since each person has unique facial structures and expressions, this creates distinct feature patterns and results in distribution shifts. This is the most challenging part that has hardly been solved in previous work~\cite{wang2024uface, gao2021sonicface}. To assess the performance of \S in this user-independent setting, we train the model using a dataset from 16 men and test it using a dataset from 4 women.

We begin by showcasing the effectiveness of the learned contrastive embeddings using the K-means model to cluster the feature embeddings. We randomly select 2,048 samples from the men's dataset (Figure~\ref{fig:train}) and the testing dataset from the 4 women's dataset (Figure~\ref{fig:test}). The K-means model successfully distinguishes the six types of expressions in both datasets, demonstrating that the learned representations are consistent and distinctive. This enables the proposed model to effectively generate positive and negative pairs for contrastive attention learning. As illustrated in the heatmap in Fig.~\ref{fig:case2}, the average accuracy is 94.9\%, with some misclassified samples for each type of expression due to variations in user expressions.

\begin{figure}[t]
\centering
\includegraphics[width=0.9\linewidth]{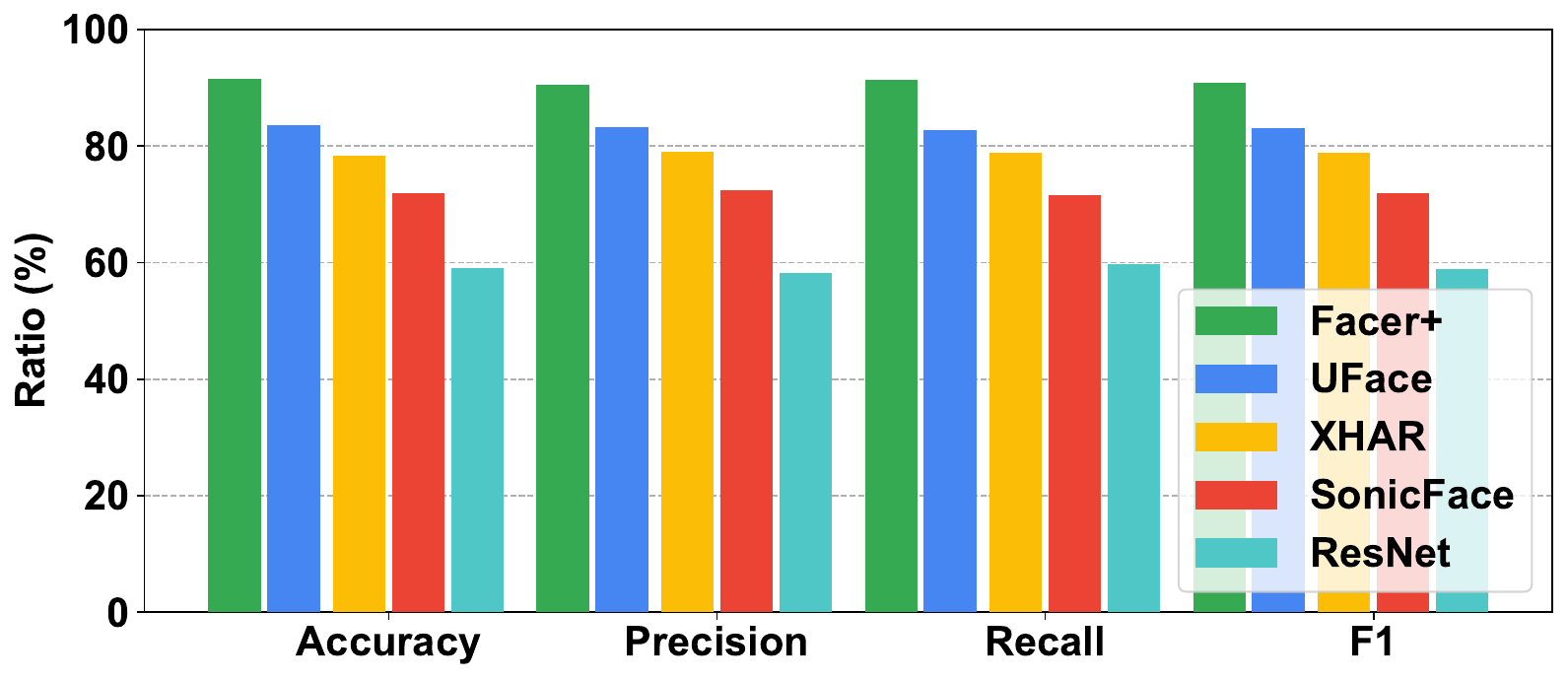}
\caption{The comparison across different models in Case 3.}
\label{fig:model_compare}
\end{figure}

\textbf{Case 3.} Next, in comparison to Case 2, we reduce the training dataset size and increase the test data size with a group of 10 independent users. As shown in Fig.~\ref{fig:case3}, the average accuracy is 91.6\%. Compared with the results in Fig.~\ref{fig:case2} and~\ref{fig:case1-1}, the performance in the Case-3 setting degrades because of the reduced training dataset, while the results of the leave-one-user-out test are slightly lower than the results from Case 2 is due to the small size of the test dataset. Furthermore, to evaluate the effectiveness of \S's domain adaptation, we compare \S with four baselines: UFacer~\cite{wang2024uface}, XHAR~\cite{zhou2020xhar}, SonicFace~\cite{gao2021sonicface}, and ResNet~\cite{he2016deep}, as illustrated in Fig.~\ref{fig:model_compare}. For instance, SonicFace uses both FMCW and pure tone signals to extract different features and employs 1D convolution for feature extraction. SonicFace can only achieve 71.9\% accuracy and 72\% F1 value.

By contrast, we treat the spectrogram of the received echoes as an image, representing the instantaneous static facial expression. Different expressions produce distinct spectrogram features, similar to how various pixels form facial expression images. Thus, we employ 2D convolution. In the user-independent Case 3 scenario, \S achieves 91.6\% accuracy and a 90.9\% F1 score as shown in Fig.~\ref{fig:model_compare}. The adversarial training-based XHAR method only attains 78.4\% accuracy and a 78.9\% F1 score. The results show the superior performance of our proposed contrastive external attention-based representation learning method, which effectively extracts robust and accurate acoustic facial expression features.

\subsection{Mask Factor Evaluation}
\textbf{Case 4 and 5.} Mask-wearing has become very common in our daily life. However, masks pose significant challenges to camera-based facial expression recognition and face recognition models~\cite{damer2020effect, wang2023masked, wang2023facer, zhang2024face}.  We examine the performance of \S when individuals wear masks. As previously mentioned, half of the dataset consists of volunteers wearing masks (mask dataset), and the other half comprises volunteers without masks (plain dataset). We train \S on the plain dataset and test it on the mask dataset, with the results shown in Figure~\ref{fig:case4}. The average accuracy is 91.8\%. Then, we train \S on the mask dataset and test it on the plain dataset, with the results presented in Figure~\ref{fig:case5}. The average accuracy is 90.7\%. As indicated, masks can affect \S's performance for expression recognition.

In particular, when trained on the dataset of masked users, \S struggles to recognize sadness, achieving only 86.9\% accuracy. In the predicted labels of ``sadness" in Case 5, there are 9\% of them that are actual ``happiness". One reason is that the fine-grained acoustic features of sadness are disturbed by the reflections from the facial mask in the mask's dataset. Thus, when the model is trained on the mask's dataset, it can hardly extract robust features of ``sadness".  Another notable finding is that \S achieves its highest accuracy of 94.7\% for the ``fear" expression when trained on either the plain dataset or the mask's dataset. Overall, under various mask conditions, \S still achieves high expression recognition accuracy, demonstrating the cross-domain adaptation capability of our proposed contrastive attention-based representation learning method.

\subsection{Ablation Study}
\subsubsection{Data Augmentation}
We evaluate the impact of data augmentation on the improvement of model performance. Taking Case 3 as an example, where we have data from 10 people as the training dataset, and data from another 10 people as the test dataset. We compare \S with two main methods (Resnet and UFace) when they are trained with data augmentation and without data augmentation. As shown in Table~\ref{tab:aug}, we can see that the model performance gets enhanced when the training data is enlarged with our designed data augmentation methods. For instance, when Facer+ is only trained with the original dataset, the average accuracy is only 85.9\%, while the average accuracy is 91.7\% when trained with the augmented dataset. Similarly, the augmented dataset can also boost the performance of other models such as Resnet and UFacer. The designed data augmentation method can increase the variety and diversity of expression sensing data across different users,  helping train the model to learn robust expression representation features. 

\begin{table}[t]
\centering
\caption{The effect of data augmentation on Case 3 Dataset.}
\label{tab:aug}
\begin{tabular}{@{}lllllll@{}}
\toprule
Model      & Anger & Disgust & Fear & Happiness & Sad  & Suprise \\ \midrule
Resnet w/o & 58.3  & 59.6    & 56.7 & 55.4      & 58.3 & 59.6    \\
Resnet w/  & 60.2  & 61.5    & 57.7 & 58.4      & 59.2 & 61.7    \\
UFacer w/o & 75.2  & 74.4    & 76.3 & 75.1      & 77.2 & 78.3    \\
UFacer w/  & 80.7  & 80.2    & 81.6 & 80.9      & 82.4 & 83.9    \\
Facer+ w/o     & 82.4  & 79.5    & 90.3 & 82.8      & 89.1 & 91.2    \\
Facer+ w/    & 89.9  & 85.6    & 96.2 & 87.4      & 94.1 & 96.7    \\ \bottomrule
\end{tabular}
\end{table}

\subsubsection{Multi-task Learning Objective}
\begin{figure}[t]
\centering
\includegraphics[width=0.9\linewidth]{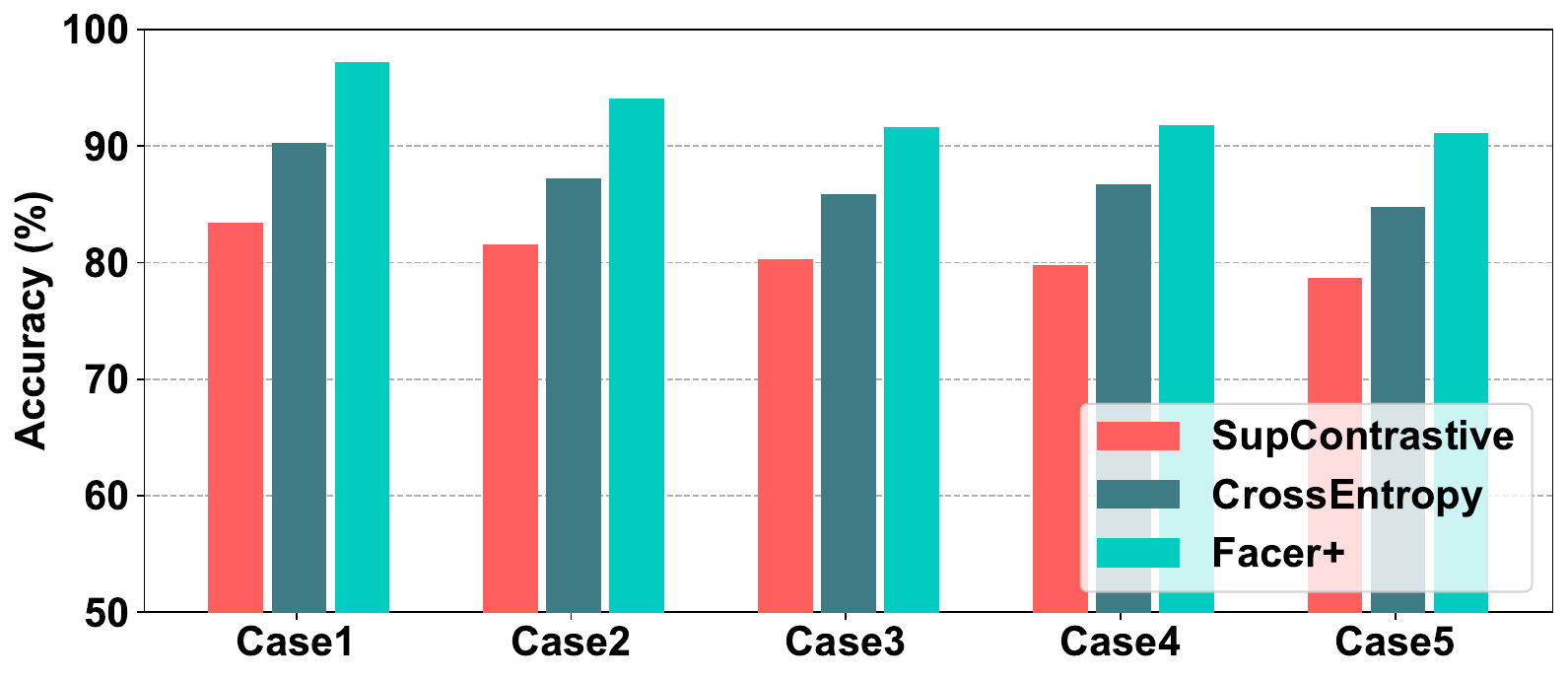}
\caption{The comparison across different learning objectives.}
\label{fig:loss_compare}
\end{figure}

We assess the effectiveness of the multi-task learning objective in \S for facial expression recognition. We compare our custom loss function in Eq.\ref{eq:loss} with the traditional CrossEntropy loss function and the supervised contrastive learning (SupContrastive) loss function. We use different learning objectives to train the model on various case datasets. As shown in Figure\ref{fig:loss_compare}, the naive SupContrastive objective fails to achieve good performance. In classification tasks with subtle and nuanced differences, such as acoustic-based expression recognition, SupContrastive underperforms struggle because it focuses on distinguishing between pairs rather than learning specific class boundaries, leading to poor performance when fine-grained distinctions are crucial. SupContrastive encourages separation between pairs rather than learning specific class prototypes. In contrast, CrossEntropy is designed for classification tasks, making it more effective in directly optimizing for class-specific decision boundaries. To achieve superior cross-domain adaptation performance, as outlined in Algorithm~\ref{alg:rep}, we designed the multi-task learning objective with Eq.~\ref{eq:lambda}. This allows \S to align feature representations in both the training and test data domains while capturing fine-grained differences necessary for expression classification.

\section{Discussion}
\label{discussion}
The performance of \S heavily relies on both the quality and quantity of acoustic sensing data. We are committed to enhancing the fairness of our facial expression recognition system by, for instance, gathering more data from female users and underrepresented groups to better capture the diversity of facial expressions. Additionally, we will explore other forms of emotional expressions, such as hand gestures. In this study, we focused primarily on scenarios where users hold the phone at a distance of 20-50 cm. In the future, we aim to examine the performance of \S when the distance between the user and the phone is greater. Such scenarios present greater challenges due to the weakening of the acoustic signal and increased reflections from nearby objects. Furthermore, we recognize the variations in phone hardware design, particularly concerning the direct path from the speaker to the microphone. Therefore, we will assess the impact of different smartphones on ultrasound signal transmission.

\section{Conclusion}
\label{conclusion}
In this work, we designed \S, a data-centric facial expression recognition system based on acoustic sensing on mobile devices. A long-standing challenge is that the distribution shift between the model training dataset and test dataset heavily impacts model inference performance. To solve this problem, we have designed a novel acoustic sensing data augmentation algorithm and contrastive attention-based learning algorithm to mitigate the distribution shift caused by many factors such as users, environment, mask, and time. In this way, \S learns robust expression features across different users in various noisy scenarios. We have conducted extensive real-world experiments, which show that \S achieves expression recognition with more than 90\% accuracy even when the users are wearing a mask.

\newpage
\balance

\bibliographystyle{IEEEtran}
\bibliography{references}

\begin{thebibliography}{10}
\providecommand{\url}[1]{#1}
\csname url@samestyle\endcsname
\providecommand{\newblock}{\relax}
\providecommand{\bibinfo}[2]{#2}
\providecommand{\BIBentrySTDinterwordspacing}{\spaceskip=0pt\relax}
\providecommand{\BIBentryALTinterwordstretchfactor}{4}
\providecommand{\BIBentryALTinterwordspacing}{\spaceskip=\fontdimen2\font plus
\BIBentryALTinterwordstretchfactor\fontdimen3\font minus
  \fontdimen4\font\relax}
\providecommand{\BIBforeignlanguage}[2]{{%
\expandafter\ifx\csname l@#1\endcsname\relax
\typeout{** WARNING: IEEEtran.bst: No hyphenation pattern has been}%
\typeout{** loaded for the language `#1'. Using the pattern for}%
\typeout{** the default language instead.}%
\else
\language=\csname l@#1\endcsname
\fi
#2}}
\providecommand{\BIBdecl}{\relax}
\BIBdecl

\bibitem{chen2021exgsense}
C.~Chen, K.~Sun, and X.~Zhang, ``Exgsense: Toward facial gesture sensing with a
  sparse near-eye sensor array,'' in \emph{Proceedings of the 20th
  International Conference on Information Processing in Sensor Networks
  (co-located with CPS-IoT Week 2021)}, 2021, pp. 222--237.

\bibitem{gao2021sonicface}
Y.~Gao, Y.~Jin, S.~Choi, J.~Li, J.~Pan, L.~Shu, C.~Zhou, and Z.~Jin,
  ``Sonicface: Tracking facial expressions using a commodity microphone
  array,'' \emph{Proceedings of the ACM on Interactive, Mobile, Wearable and
  Ubiquitous Technologies}, vol.~5, no.~4, pp. 1--33, 2021.

\bibitem{krumhuber2023role}
E.~G. Krumhuber, L.~I. Skora, H.~C. Hill, and K.~Lander, ``The role of facial
  movements in emotion recognition,'' \emph{Nature Reviews Psychology}, vol.~2,
  no.~5, pp. 283--296, 2023.

\bibitem{nwe2001speech}
T.~L. Nwe, F.~S. Wei, and L.~C. De~Silva, ``Speech based emotion
  classification,'' in \emph{Proceedings of IEEE Region 10 International
  Conference on Electrical and Electronic Technology. TENCON 2001 (Cat. No.
  01CH37239)}, vol.~1.\hskip 1em plus 0.5em minus 0.4em\relax IEEE, 2001, pp.
  297--301.

\bibitem{swain2018databases}
M.~Swain, A.~Routray, and P.~Kabisatpathy, ``Databases, features and
  classifiers for speech emotion recognition: a review,'' \emph{International
  Journal of Speech Technology}, vol.~21, pp. 93--120, 2018.

\bibitem{wani2021comprehensive}
T.~M. Wani, T.~S. Gunawan, S.~A.~A. Qadri, M.~Kartiwi, and E.~Ambikairajah, ``A
  comprehensive review of speech emotion recognition systems,'' \emph{IEEE
  access}, vol.~9, pp. 47\,795--47\,814, 2021.

\bibitem{zhao2016emotion}
M.~Zhao, F.~Adib, and D.~Katabi, ``Emotion recognition using wireless
  signals,'' in \emph{Proceedings of the 22nd annual international conference
  on mobile computing and networking}, 2016, pp. 95--108.

\bibitem{saganowski2022emotion}
S.~Saganowski, B.~Perz, A.~Polak, and P.~Kazienko, ``Emotion recognition for
  everyday life using physiological signals from wearables: A systematic
  literature review,'' \emph{IEEE Transactions on Affective Computing}, 2022.

\bibitem{zhang2020emotion}
X.~Zhang, J.~Liu, J.~Shen, S.~Li, K.~Hou, B.~Hu, J.~Gao, and T.~Zhang,
  ``Emotion recognition from multimodal physiological signals using a
  regularized deep fusion of kernel machine,'' \emph{IEEE transactions on
  cybernetics}, vol.~51, no.~9, pp. 4386--4399, 2020.

\bibitem{li2020deep}
S.~Li and W.~Deng, ``Deep facial expression recognition: A survey,'' \emph{IEEE
  transactions on affective computing}, 2020.

\bibitem{ekmann1973universal}
P.~Ekmann, ``Universal facial expressions in emotion,'' \emph{Studia
  Psychologica}, vol.~15, no.~2, p. 140, 1973.

\bibitem{clark2020facial}
E.~A. Clark, J.~Kessinger, S.~E. Duncan, M.~A. Bell, J.~Lahne, D.~L. Gallagher,
  and S.~F. O'Keefe, ``The facial action coding system for characterization of
  human affective response to consumer product-based stimuli: a systematic
  review,'' \emph{Frontiers in psychology}, vol.~11, p. 920, 2020.

\bibitem{cao2013facewarehouse}
C.~Cao, Y.~Weng, S.~Zhou, Y.~Tong, and K.~Zhou, ``Facewarehouse: A 3d facial
  expression database for visual computing,'' \emph{IEEE Transactions on
  Visualization and Computer Graphics}, vol.~20, no.~3, pp. 413--425, 2013.

\bibitem{patil2016real}
J.~V. Patil and P.~Bailke, ``Real time facial expression recognition using
  realsense camera and ann,'' in \emph{2016 International Conference on
  Inventive Computation Technologies (ICICT)}, vol.~2.\hskip 1em plus 0.5em
  minus 0.4em\relax IEEE, 2016, pp. 1--6.

\bibitem{shen2016facial}
T.-W. Shen, H.~Fu, J.~Chen, W.~Yu, C.~Lau, W.~Lo, and Z.~Chi, ``Facial
  expression recognition using depth map estimation of light field camera,'' in
  \emph{2016 IEEE International Conference on Signal Processing, Communications
  and Computing (ICSPCC)}.\hskip 1em plus 0.5em minus 0.4em\relax IEEE, 2016,
  pp. 1--4.

\bibitem{gu2020wife}
Y.~Gu, X.~Zhang, Z.~Liu, and F.~Ren, ``Wife: Wifi and vision based intelligent
  facial-gesture emotion recognition,'' \emph{arXiv preprint arXiv:2004.09889},
  2020.

\bibitem{chen2020wiface}
Y.~Chen, R.~Ou, Z.~Li, and K.~Wu, ``Wiface: facial expression recognition using
  wi-fi signals,'' \emph{IEEE Transactions on Mobile Computing}, vol.~21,
  no.~1, pp. 378--391, 2020.

\bibitem{choi2022ppgface}
S.~Choi, Y.~Gao, Y.~Jin, S.~j. Kim, J.~Li, W.~Xu, and Z.~Jin, ``Ppgface: Like
  what you are watching? earphones can" feel" your facial expressions,''
  \emph{Proceedings of the ACM on Interactive, Mobile, Wearable and Ubiquitous
  Technologies}, vol.~6, no.~2, pp. 1--32, 2022.

\bibitem{wang2023facer}
G.~Wang, Q.~Yan, S.~Patrarungrong, J.~Wang, and H.~Zeng, ``Facer: Contrastive
  attention based expression recognition via smartphone earpiece speaker,'' in
  \emph{IEEE INFOCOM 2023-IEEE Conference on Computer Communications}.\hskip
  1em plus 0.5em minus 0.4em\relax IEEE, 2023, pp. 1--10.

\bibitem{wang2024uface}
S.~Wang, L.~Zhong, Y.~Fu, L.~Chen, J.~Ren, and Y.~Zhang, ``Uface: Your
  smartphone can" hear" your facial expression!'' \emph{Proceedings of the ACM
  on Interactive, Mobile, Wearable and Ubiquitous Technologies}, vol.~8, no.~1,
  pp. 1--27, 2024.

\bibitem{aneja2016modeling}
D.~Aneja, A.~Colburn, G.~Faigin, L.~Shapiro, and B.~Mones, ``Modeling stylized
  character expressions via deep learning,'' in \emph{Asian Conference on
  Computer Vision}.\hskip 1em plus 0.5em minus 0.4em\relax Springer, 2016, pp.
  136--153.

\bibitem{gruebler2014design}
A.~Gruebler and K.~Suzuki, ``Design of a wearable device for reading positive
  expressions from facial emg signals,'' \emph{IEEE Transactions on affective
  computing}, vol.~5, no.~3, pp. 227--237, 2014.

\bibitem{rommel2012heart}
D.~Rommel, J.~Nandrino, M.~Jeanne, R.~Logier \emph{et~al.}, ``Heart rate
  variability analysis as an index of emotion regulation processes: interest of
  the analgesia nociception index (ani).'' in \emph{2012 Annual international
  conference of the IEEE engineering in medicine and biology society}.\hskip
  1em plus 0.5em minus 0.4em\relax IEEE, 2012, pp. 3432--3435.

\bibitem{qian2018acousticcardiogram}
K.~Qian, C.~Wu, F.~Xiao, Y.~Zheng, Y.~Zhang, Z.~Yang, and Y.~Liu,
  ``Acousticcardiogram: Monitoring heartbeats using acoustic signals on smart
  devices,'' in \emph{IEEE INFOCOM 2018-IEEE conference on computer
  communications}.\hskip 1em plus 0.5em minus 0.4em\relax IEEE, 2018, pp.
  1574--1582.

\bibitem{fleureau2013affective}
J.~Fleureau, P.~Guillotel, and I.~Orlac, ``Affective benchmarking of movies
  based on the physiological responses of a real audience,'' in \emph{2013
  Humaine Association Conference on Affective Computing and Intelligent
  Interaction}.\hskip 1em plus 0.5em minus 0.4em\relax IEEE, 2013, pp. 73--78.

\bibitem{verma2021expressear}
D.~Verma, S.~Bhalla, D.~Sahnan, J.~Shukla, and A.~Parnami, ``Expressear:
  Sensing fine-grained facial expressions with earables,'' \emph{Proceedings of
  the ACM on Interactive, Mobile, Wearable and Ubiquitous Technologies},
  vol.~5, no.~3, pp. 1--28, 2021.

\bibitem{chen2021neckface}
T.~Chen, Y.~Li, S.~Tao, H.~Lim, M.~Sakashita, R.~Zhang, F.~Guimbretiere, and
  C.~Zhang, ``Neckface: Continuously tracking full facial expressions on
  neck-mounted wearables,'' \emph{Proceedings of the ACM on Interactive,
  Mobile, Wearable and Ubiquitous Technologies}, vol.~5, no.~2, pp. 1--31,
  2021.

\bibitem{songfacelistener2022}
X.~Song, K.~Huang, and W.~Gao, ``Facelistener: Recognizing human facial
  expressions via acoustic sensing on commodity headphones,'' in \emph{21st
  ACM/IEEE International Conference on Information Processing in Sensor
  Networks (IPSN)}, 2022.

\bibitem{wang2018socialite}
G.~Wang, L.~Zhang, Z.~Yang, and X.-Y. Li, ``Socialite: Social activity mining
  and friend auto-labeling,'' in \emph{2018 IEEE 37th International Performance
  Computing and Communications Conference (IPCCC)}.\hskip 1em plus 0.5em minus
  0.4em\relax IEEE, 2018, pp. 1--8.

\bibitem{han2019shad}
F.~Han, L.~Zhang, X.~You, G.~Wang, and X.-Y. Li, ``Shad: Privacy-friendly
  shared activity detection and data sharing,'' in \emph{2019 IEEE 16th
  International Conference on Mobile Ad Hoc and Sensor Systems (MASS)}.\hskip
  1em plus 0.5em minus 0.4em\relax IEEE, 2019, pp. 109--117.

\bibitem{zheng2019zero}
Y.~Zheng, Y.~Zhang, K.~Qian, G.~Zhang, Y.~Liu, C.~Wu, and Z.~Yang,
  ``Zero-effort cross-domain gesture recognition with wi-fi,'' in
  \emph{Proceedings of the 17th Annual International Conference on Mobile
  Systems, Applications, and Services}, 2019, pp. 313--325.

\bibitem{li2020wihf}
C.~Li, M.~Liu, and Z.~Cao, ``Wihf: enable user identified gesture recognition
  with wifi,'' in \emph{IEEE INFOCOM 2020-IEEE Conference on Computer
  Communications}.\hskip 1em plus 0.5em minus 0.4em\relax IEEE, 2020, pp.
  586--595.

\bibitem{li2021deep}
C.~Li, Z.~Cao, and Y.~Liu, ``Deep ai enabled ubiquitous wireless sensing: A
  survey,'' \emph{ACM Computing Surveys (CSUR)}, vol.~54, no.~2, pp. 1--35,
  2021.

\bibitem{rostaminia2019w}
S.~Rostaminia, A.~Lamson, S.~Maji, T.~Rahman, and D.~Ganesan, ``W! nce:
  Unobtrusive sensing of upper facial action units with eog-based eyewear,''
  \emph{Proceedings of the ACM on Interactive, Mobile, Wearable and Ubiquitous
  Technologies}, vol.~3, no.~1, pp. 1--26, 2019.

\bibitem{hof2020face}
E.~Hof, A.~Sanderovich, M.~Salama, and E.~Hemo, ``Face verification using
  mmwave radar sensor,'' in \emph{2020 International Conference on Artificial
  Intelligence in Information and Communication (ICAIIC)}.\hskip 1em plus 0.5em
  minus 0.4em\relax IEEE, 2020, pp. 320--324.

\bibitem{xu2019breathlistener}
X.~Xu, J.~Yu, Y.~Chen, Y.~Zhu, L.~Kong, and M.~Li, ``Breathlistener:
  Fine-grained breathing monitoring in driving environments utilizing acoustic
  signals,'' in \emph{Proceedings of the 17th Annual International Conference
  on Mobile Systems, Applications, and Services}, 2019, pp. 54--66.

\bibitem{zhou2018echoprint}
B.~Zhou, J.~Lohokare, R.~Gao, and F.~Ye, ``Echoprint: Two-factor authentication
  using acoustics and vision on smartphones,'' in \emph{Proceedings of the 24th
  Annual International Conference on Mobile Computing and Networking}, 2018,
  pp. 321--336.

\bibitem{gao2022mom}
Z.~Gao, A.~Li, D.~Li, J.~Liu, J.~Xiong, Y.~Wang, B.~Li, and Y.~Chen, ``Mom:
  Microphone based 3d orientation measurement,'' in \emph{2022 21st ACM/IEEE
  International Conference on Information Processing in Sensor Networks
  (IPSN)}.\hskip 1em plus 0.5em minus 0.4em\relax IEEE, 2022, pp. 132--144.

\bibitem{xie2022teethpass}
Y.~Xie, F.~Li, Y.~Wu, H.~Chen, Z.~Zhao, and Y.~Wang, ``Teethpass: Dental
  occlusion-based user authentication via in-ear acoustic sensing,'' in
  \emph{IEEE INFOCOM 2022-IEEE Conference on Computer Communications}.\hskip
  1em plus 0.5em minus 0.4em\relax IEEE, 2022, pp. 1789--1798.

\bibitem{li2022lasense}
D.~Li, J.~Liu, S.~I. Lee, and J.~Xiong, ``Lasense: Pushing the limits of
  fine-grained activity sensing using acoustic signals,'' \emph{Proceedings of
  the ACM on Interactive, Mobile, Wearable and Ubiquitous Technologies},
  vol.~6, no.~1, pp. 1--27, 2022.

\bibitem{mao2019rnn}
W.~Mao, M.~Wang, W.~Sun, L.~Qiu, S.~Pradhan, and Y.-C. Chen, ``Rnn-based room
  scale hand motion tracking,'' in \emph{The 25th Annual International
  Conference on Mobile Computing and Networking}, 2019, pp. 1--16.

\bibitem{gao2020echowhisper}
Y.~Gao, Y.~Jin, J.~Li, S.~Choi, and Z.~Jin, ``Echowhisper: Exploring an
  acoustic-based silent speech interface for smartphone users,''
  \emph{Proceedings of the ACM on Interactive, Mobile, Wearable and Ubiquitous
  Technologies}, vol.~4, no.~3, pp. 1--27, 2020.

\bibitem{zhang2021soundlip}
Q.~Zhang, D.~Wang, R.~Zhao, and Y.~Yu, ``Soundlip: Enabling word and
  sentence-level lip interaction for smart devices,'' \emph{Proceedings of the
  ACM on Interactive, Mobile, Wearable and Ubiquitous Technologies}, vol.~5,
  no.~1, pp. 1--28, 2021.

\bibitem{iravantchi2019beamband}
Y.~Iravantchi, M.~Goel, and C.~Harrison, ``Beamband: Hand gesture sensing with
  ultrasonic beamforming,'' in \emph{Proceedings of the 2019 CHI Conference on
  Human Factors in Computing Systems}, 2019, pp. 1--10.

\bibitem{li2022eario}
K.~Li, R.~Zhang, B.~Liang, F.~Guimbreti{\`e}re, and C.~Zhang, ``Eario: A
  low-power acoustic sensing earable for continuously tracking detailed facial
  movements,'' \emph{Proceedings of the ACM on Interactive, Mobile, Wearable
  and Ubiquitous Technologies}, vol.~6, no.~2, pp. 1--24, 2022.

\bibitem{zhang2024face}
Y.~Zhang, P.~Tong, S.~Li, Y.~Xie, and M.~Li, ``Face recognition in harsh
  conditions: An acoustic based approach,'' in \emph{Proceedings of the 22nd
  Annual International Conference on Mobile Systems, Applications and
  Services}, 2024, pp. 1--14.

\bibitem{xie2021acoustic}
W.~Xie, Q.~Zhang, and J.~Zhang, ``Acoustic-based upper facial action
  recognition for smart eyewear,'' \emph{Proceedings of the ACM on Interactive,
  Mobile, Wearable and Ubiquitous Technologies}, vol.~5, no.~2, pp. 1--28,
  2021.

\bibitem{wei2018person}
L.~Wei, S.~Zhang, W.~Gao, and Q.~Tian, ``Person transfer gan to bridge domain
  gap for person re-identification,'' in \emph{Proceedings of the IEEE
  conference on computer vision and pattern recognition}, 2018, pp. 79--88.

\bibitem{zhou2020xhar}
Z.~Zhou, Y.~Zhang, X.~Yu, P.~Yang, X.-Y. Li, J.~Zhao, and H.~Zhou, ``Xhar: Deep
  domain adaptation for human activity recognition with smart devices,'' in
  \emph{2020 17th Annual IEEE International Conference on Sensing,
  Communication, and Networking (SECON)}.\hskip 1em plus 0.5em minus
  0.4em\relax IEEE, 2020, pp. 1--9.

\bibitem{vaswani2017attention}
A.~Vaswani, N.~Shazeer, N.~Parmar, J.~Uszkoreit, L.~Jones, A.~N. Gomez,
  {\L}.~Kaiser, and I.~Polosukhin, ``Attention is all you need,''
  \emph{Advances in neural information processing systems}, vol.~30, 2017.

\bibitem{khosla2020supervised}
P.~Khosla, P.~Teterwak, C.~Wang, A.~Sarna, Y.~Tian, P.~Isola, A.~Maschinot,
  C.~Liu, and D.~Krishnan, ``Supervised contrastive learning,'' \emph{Advances
  in Neural Information Processing Systems}, vol.~33, pp. 18\,661--18\,673,
  2020.

\bibitem{mao2018aim}
W.~Mao, M.~Wang, and L.~Qiu, ``Aim: Acoustic imaging on a mobile,'' in
  \emph{Proceedings of the 16th Annual International Conference on Mobile
  Systems, Applications, and Services}, 2018, pp. 468--481.

\bibitem{caron2018deep}
M.~Caron, P.~Bojanowski, A.~Joulin, and M.~Douze, ``Deep clustering for
  unsupervised learning of visual features,'' in \emph{Proceedings of the
  European conference on computer vision (ECCV)}, 2018, pp. 132--149.

\bibitem{yang2021cade}
L.~Yang, W.~Guo, Q.~Hao, A.~Ciptadi, A.~Ahmadzadeh, X.~Xing, and G.~Wang,
  ``$\{$CADE$\}$: Detecting and explaining concept drift samples for security
  applications,'' in \emph{30th USENIX Security Symposium (USENIX Security
  21)}, 2021, pp. 2327--2344.

\bibitem{leys2013detecting}
C.~Leys, C.~Ley, O.~Klein, P.~Bernard, and L.~Licata, ``Detecting outliers: Do
  not use standard deviation around the mean, use absolute deviation around the
  median,'' \emph{Journal of experimental social psychology}, vol.~49, no.~4,
  pp. 764--766, 2013.

\bibitem{guo2021beyond}
M.-H. Guo, Z.-N. Liu, T.-J. Mu, and S.-M. Hu, ``Beyond self-attention: External
  attention using two linear layers for visual tasks,'' \emph{arXiv preprint
  arXiv:2105.02358}, 2021.

\bibitem{tung2018cross}
Y.-C. Tung, D.~Bui, and K.~G. Shin, ``Cross-platform support for rapid
  development of mobile acoustic sensing applications,'' in \emph{Proceedings
  of the 16th Annual International Conference on Mobile Systems, Applications,
  and Services}, 2018, pp. 455--467.

\bibitem{chen2020chaperone}
J.~Chen, U.~Hengartner, H.~Khan, and M.~Mannan, ``Chaperone: Real-time locking
  and loss prevention for smartphones,'' in \emph{29th USENIX Security
  Symposium (USENIX Security 20)}, 2020, pp. 325--342.

\bibitem{he2016deep}
K.~He, X.~Zhang, S.~Ren, and J.~Sun, ``Deep residual learning for image
  recognition,'' in \emph{Proceedings of the IEEE conference on computer vision
  and pattern recognition}, 2016, pp. 770--778.

\bibitem{scikit}
R.~Nakano, ``Scikit-plot,'' \url{https://github.com/reiinakano/scikit-plot},
  2017.

\bibitem{damer2020effect}
N.~Damer, J.~H. Grebe, C.~Chen, F.~Boutros, F.~Kirchbuchner, and A.~Kuijper,
  ``The effect of wearing a mask on face recognition performance: an
  exploratory study,'' in \emph{2020 International Conference of the Biometrics
  Special Interest Group (BIOSIG)}.\hskip 1em plus 0.5em minus 0.4em\relax
  IEEE, 2020, pp. 1--6.

\bibitem{wang2023masked}
Z.~Wang, B.~Huang, G.~Wang, P.~Yi, and K.~Jiang, ``Masked face recognition
  dataset and application,'' \emph{IEEE Transactions on Biometrics, Behavior,
  and Identity Science}, vol.~5, no.~2, pp. 298--304, 2023.

\end{thebibliography}

 





\end{document}